\documentclass[letterpaper]{article} % DO NOT CHANGE THIS
\usepackage{aaai23}  % DO NOT CHANGE THIS
\usepackage{times}  % DO NOT CHANGE THIS
\usepackage{helvet}  % DO NOT CHANGE THIS
\usepackage{courier}  % DO NOT CHANGE THIS
\usepackage[hyphens]{url}  % DO NOT CHANGE THIS
\usepackage{graphicx} % DO NOT CHANGE THIS
\usepackage{natbib}  % DO NOT CHANGE THIS AND DO NOT ADD ANY OPTIONS TO IT
\usepackage{caption} % DO NOT CHANGE THIS AND DO NOT ADD ANY OPTIONS TO IT
\usepackage{algorithm}
\usepackage{algorithmic}

\usepackage{newfloat}
\usepackage{listings}
\DeclareCaptionStyle{ruled}{labelfont=normalfont,labelsep=colon,strut=off} % DO NOT CHANGE THIS
\floatstyle{ruled}
\newfloat{listing}{tb}{lst}{}
\floatname{listing}{Listing}
\usepackage{adjustbox, multirow, paralist, amsmath, amssymb}

\title{Boosting Video Super Resolution with Patch-Based Temporal Redundancy Optimization}

\author{
    Yuhao Huang\textsuperscript{\rm 1}~\equalcontrib,
    Hang Dong\textsuperscript{\rm 2}~\equalcontrib~\thanks{Corresponding author.},
    Jinshan Pan\textsuperscript{\rm 3},
    Chao Zhu\textsuperscript{\rm 1},
    Boyang Liang\textsuperscript{\rm 1},
    Yu Guo\textsuperscript{\rm 1},
    Ding Liu\textsuperscript{\rm 2},
    Lean Fu\textsuperscript{\rm 2},
    Fei Wang\textsuperscript{\rm 1}
}
\affiliations{
    \textsuperscript{\rm 1}Xi'an Jiaotong University\\
    \textsuperscript{\rm 2}ByteDance Intelligent Creation Lab\\
    \textsuperscript{\rm 3}Nanjing University of Science and Technology\\

    hyhsimon@gmail.com
}

\usepackage{bibentry}
\begin{document}

\maketitle

\begin{abstract}
The success of existing video super-resolution (VSR) algorithms stems mainly exploiting the temporal information from the neighboring frames.
However, none of these methods have discussed the influence of the temporal redundancy in the patches with stationary objects and background and usually use all the information in the adjacent frames without any discrimination.
In this paper, we observe that the temporal redundancy will bring adverse effect to the information propagation, which limits the performance of the most existing VSR methods and causes the severe generalization problem.
Motivated by this observation, we aim to improve existing VSR algorithms by handling the temporal redundancy patches in an optimized manner.
We develop two simple yet effective plug-and-play methods to improve the performance and the generalization ability of existing local and non-local propagation-based VSR algorithms on widely-used public videos. 
For more comprehensive evaluating the robustness and performance of existing VSR algorithms, we also collect a new dataset which contains a variety of public videos as testing set.
Extensive evaluations show that the proposed methods can significantly improve the performance and the generalization ability of existing VSR methods on the collected videos from wild scenarios while maintain their performance on existing commonly used datasets.
The code is available at \emph{https://github.com/HYHsimon/Boosted-VSR}.
\end{abstract}

%%%%%%%%% BODY TEXT
\section{Introduction}
\label{sec:1}

% VSR的背景
Video Super-Resolution (VSR) aims to reconstruct a high-resolution visual-pleasing video from a low-resolution one.
Recent years have witnessed significant advances due to the use of deep convolutional neural networks (CNNs). 
As more frames are used, VSR methods achieve better performance than the single image SR methods~\cite{dbpn,rcan,san,IGNN,nlsa,ipt} on existing VSR datasets (e.g., REDS~\cite{REDS}, Vid4~\cite{Vid4}, Vimeo-90K~\cite{Vimeo}). 
However, the VSR task introduces another challenging problem, i.e., how to effectively exploit the temporal information for better results.

% VSR method的成功是由于temporal信息的利用
To solve this problem, most existing deep learning-based methods usually employ optical flow, deformable convolution networks, and recurrent neural networks to explore useful information from adjacent frames for better high-resolution video restoration.
Existing deep learning-based VSR methods can be roughly categorized into local propagation-based~(e.g., EDVR) and non-local propagation-based~(e.g., BasicVSR) methods according to the propagation scheme of the input frames.
The success of existing VSR stems mainly exploiting the temporal information from the neighboring frames through propagation.

%============= Fig: Temporal Redundancy Validation =============%
\begin{figure*}[!t] 
  \centering
  \includegraphics[width=0.85\linewidth]{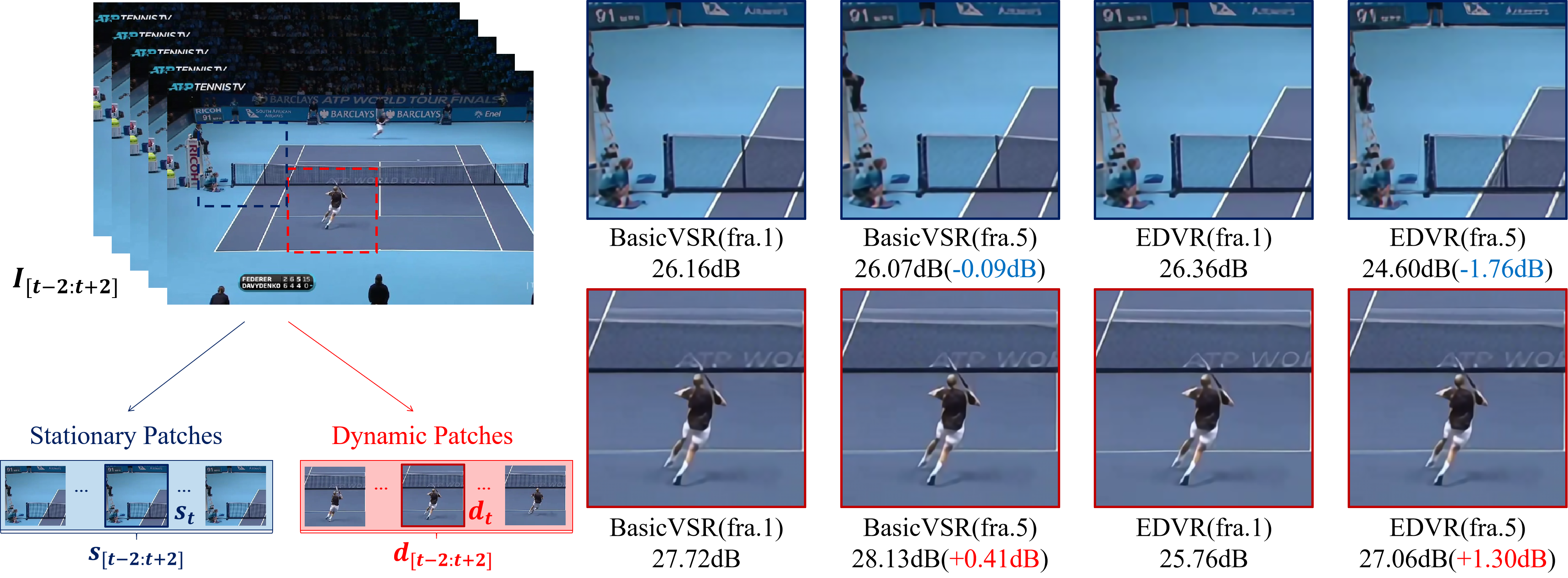}
  \caption{
  \textbf{Effect of the temporal redundancy in stationary objects and background.} 
  Since the EDVR can only received five frames as inputs, we modify the original EDVR to adapt the single frame input~(more details can be find in Sec.~\ref{sec:4.1}).
  Both the EDVR and BasicVSR are trained on the REDS dataset.
  fra.N means method takes N frames as input.
  }
\label{fig:Validation}
\end{figure*}

% 提出temporal redundancy
Meanwhile, we note that the neighboring frames also contains similar contents~(i.e., temporal redundancy) in the patches with the stationary objects and background. 
If these temporal redundancy contents dominate the propagation process, they will not facilitate the VSR problem as no additional useful information is introduced from the temporal domain.
However, most existing methods usually use all the information from adjacent frames without any discrimination. Therefore, the temporal redundancy are likely to be involved in the high-resolution frame reconstruction process.

% fig.1观察实验
In this paper, we find that the temporal redundancy in stationary objects and background interfere with the high-resolution frame reconstruction if they are not specially handled. 
As shown in Figure~\ref{fig:Validation}, we select one patch sequence with stationary objects and background $s_{[t-2:t+2]}$ and one patch sequence with dynamic scene $d_{[t-2:t+2]}$ from input frames $I_{[t-2:t+2]}$ and super-resolve them with two typical VSR networks in the local~(EDVR~\cite{wang2019edvr}) and non-local~(BasicVSR~\cite{basicvsr}) propagation-based methods.
To evaluate the benefit of neighboring frames, we also super-resolve the reference patches~($s_{t}$ and $d_{t}$) with two single frame counterparts of these two methods for comparisons. 
The super-resolved results of these two patch sequences are shown in the right side of Figure~\ref{fig:Validation}.
As expected, by exploiting the temporal information from the neighboring patches, both networks can achieve better results in the dynamic patch.
In the meantime, due to the existence of temporal redundancy contents, the single frame counterparts outperform the VSR networks in the patch with stationary objects and background.
The inconsistent performance of the VSR networks on two patch sequences demonstrates that the temporal redundancy may bring adverse effect on the VSR problem and patches with stationary background and dynamic objects should be handled separately.
% 对于adverse effect的分析
VSR networks exploit the useful subpixel temporal information though alignment from the neighboring patches with dynamic objects. However, pixels may still change due to noisy and information loss during encoding and decoding in the patches with stationary background, which will be regarded as the useful temporal information by the alignment module and bring adverse effect on the VSR problem.

% 提出Boosted EDVR
To overcome this problem, we try to handle the temporal redundancy patches in an optimized manner and develop two simple yet effective plug-and-play methods to improve the performance and the generalization ability of existing VSR algorithms.
Our work is motivated by two observations: the temporal redundancy content is universal on different types of videos and the single frame super-resolution is more suitable for handling patches with temporal redundancy contents.
This inspired us to propose a new VSR pipeline with temporal redundancy detection module for local propagation-based methods and deploy it to the original EDVR, namely \textbf{Boosted EDVR}.
Specifically, the proposed pipeline first decomposes the input frames into overlapping patches and super-resolve the detected patches with a fine-tuned EDVR model for single frame~(EDVR-1F). 
Since the EDVR-1F is more suitable for super-resolving patches with temporal redundancy and has lower computational cost than the original EDVR, the \textbf{Boosted EDVR} could improve the performance and accelerate the inference time simultaneously.

% 提出Boosted BasicVSR
Moreover, we also optimize the non-local propagation-based VSR methods in a different way based on another observation:
% non-local propagation-based VSR methods的局限性
As for the non-local propagation-based VSR methods, one frame may strongly affect the next adjacent frame, but its influence is quickly lost after few time steps.
Therefore, the temporal redundancy in the patch sequences will hinder the propagation of the hidden states since the temporal information in the distant frame may be gradually vanished by the temporal redundancy from neighboring patches.
To improve the effectiveness of the hidden states propagation in the presence of temporal redundancy contents, we propose a patch-based dynamic propagation~(PDP) scheme to better accumulate and exploit the long-term information.
Unlike existing propagation schemes, where the information is sequentially propagated frame-by-frame, the proposed patch-based dynamic propagation can directly propagate the long-term information to the current frame in a patch-wise way without accumulating useless redundancy.
We deploy this propagation scheme to BasicVSR, namely \textbf{Boosted BasicVSR}, and largely improve the performance and the generalization ability without any training process.     

% 提出DTVIT dataset
In addition, we also collect a new testing dataset which contains a variety of public videos to comprehensively evaluate the robustness and performance of VSR algorithms.
More specifically, the collected testing dataset contains videos from live streaming, TV program, sports live, movie and television, surveillance camera, advertisement, and some first person videos captured with irregular trajectories.
We believe that the new dataset is suitable for evaluating the importance of temporal redundancy and can enrich the video types of the existing datasets.

% 贡献
The contributions of this work are summarized as follows:
  \begin{compactitem}
  \item In this paper, we find that the temporal redundancy is universal in public videos and will limit the potential of the existing VSR methods.
  To the best of our knowledge, this is the first work to investigate the influence of the temporal redundancy in the VSR task.
  \item We develop two plug-and-play methods for both the local and non-local propagation-based VSR methods, which can optimize the super-resolving process for the temporal redundancy patches and save computational cost.
  \item We collect a dataset with a variety of public videos to enrich the existing datasets. 
  Extensive evaluations demonstrate that the proposed methods can largely improve the performance and the generalization ability of existing VSR algorithms.
  \end{compactitem}

\section{Related Works}
\label{sec:2}

% Sliding Widow方法的发展和局限性
Most existing VSR algorithms~~\cite{chan2020understanding,isobe2020video,huang2017video,huang2015bidirectional,VESPCN,SPMC,DUF,PFVSR,TDAN,wang2019edvr,RSDN,RRN,lai2017deep,kingma2014adam,kim2018spatio,jo2018deep} focus on improving the motion compensation and frame aggregation modules to better exploit temporal information.
In VESPCN~\cite{VESPCN}, a real-time deep motion compensation module is proposed for frames registration. 
SPMC~\cite{SPMC} further improve the process by proposing a sub-pixel motion compensation (SPMC) strategy, which is validated by the physical imaging model.
Since optical flow estimation is a challenging task in dynamic scenes, some recent works adopt implicit alignment without the optical flow estimation process.
EDVR and TDAN~\cite{TDAN} both adopt deformable convolutions (DCNs~\cite{DCN}) to align the features of the neighboring frames in a multi-scale architecture.
In DUF~\cite{DUF}, a novel learned dynamic upsampling filter is proposed to exploiting the spatio-temporal of each pixel without explicit motion compensation.
Although these sliding-window frames can achieve favorable results, none of them discuss the effect of the temporal redundancy, which leads to sub-optimal results and causes unnecessary consumption.

% RNN方法的发展和局限性
Since the RNN~(non-local propagation-based) architecture has been validated to be effective in processing the time sequence signals, it is also applied in the some video super-resolution tasks. 
FRVSR~\cite{FRVSR} first proposes a recurrent network to super-resolve the low resolution video by leveraging the HR output from last iteration.
Since the propagation is one of the most influential components in non-local propagation-based VSR algorithms, subsequent methods propose new propagation schemes to improve the information-flow of the hidden states. 
RRN~\cite{RRN}  proposes a new recurrent residual block to solve the gradient vanish problem and preserve the texture information over long periods.
Recently, BasicVSR and BasicVSR++~\cite{basicvsr++} achieves SotA performance on all the existing datasets by adopting a bidirectional propagation coupled with optical flow-based and deformable-based feature alignments.
Despite the distinguished performance, the information in BasicVSR and BasicVSR++ are still sequentially propagated frame-by-frame which is not optimal when temporal redundancy patches exist.
The most similar work to our paper is RSDN~\cite{RSDN}, where a spatially variant hidden state adaptation module is proposed to only propagate the similar information in previous frames to the current frame at each position.
However, this strategy bring serious adverse effects when handling the video with temporal redundancy, since the useful information in the long-term frames will be totally replaced by the temporal redundancy contents.

\section{Observations on Temporal Redundancy}
\label{sec:3}
\subsection{The DTVIT Dataset}
\label{sec:3.1}

% 为什么提出这个数据集，一现在数据集过于单一， 二TR在公共视频中是大量存在的
Currently, most VSR datasets are first-person videos, which contains only dynamic scenes due to consistent movement.
However, there are a variety of videos with irregular movement in public videos.
To better investigate temporal redundancy and its influence, 
We collected a Diverse Types Videos with Irregular Trajectories~(DTVIT) Dataset.
More specifically, we collect 96 videos with high-quality and high-resolution as ground-truth from the internet.
To ensure the diversity of the datasets, the collected videos include live streaming, TV program, sports live, movie and television, surveillance camera, and advertisement.
Besides, to further increase the quantity and diversity of the collected dataset, 
we also additionally capture 12 first-person videos with irregular trajectories (using iPhone 12 with DJI stabilizer). More details can be find in the supplementary.
Then, we randomly select ten videos from DTVIT dataset as the validation set and try to investigate the influence of temporal redundancy based on it.

\subsection{Temporal Redundancy in Videos}
\label{sec:3.2}
{\flushleft \bf Observation 1:}
\emph{The temporal redundancy contents is universal in widely-used public videos.}

% ob1: 公共视频中TR的普遍性
As temporal redundancy occurs in the stationary objects and background, we conduct a statistical analysis on the sliced patches of the validation set to determine the ratio of the patch sequence with stationary objects and background.
Here, based on the input length of most local propagation-based VSR algorithms,
we define the five neighboring patches, where the PSNR of each neighboring patch is higher than 35, as a patch sequence with stationary objects and background.
Based on the definition above,
there are 69.92\% patch sequences in the validation set can be discriminated as stationary.
Even we extend the length of the patch sequence to 11 patches, 
there are still 64.79\% patch sequences can be treated as stationary.
These statistic results demonstrate that the patch sequence with stationary objects and background, as well as the temporal redundancy, is universal in widely-used public videos.
For convenience, the patch sequences with stationary objects and background are denoted as the Type A sequences,
while the dynamic patch sequences are denoted as the Type B sequences.

%============= Tab: Observation 2 =============%
\begin{table}[!t] 
\centering
\caption{{\bf Performance of EDVR-1f and two input types of EDVR-5f.} Type A and Type B sequences refer to the stationary and dynamic sequences.}
\begin{adjustbox}{width=0.7\linewidth}
\begin{tabular}{c c c}
    \hline
    {\bf Models}       & Type A sequences   & Type B sequences\\
    \hline
    EDVR-1F                  & {\bf39.20dB}	  & 38.01dB            \\
    EDVR-5f(original)        & 37.81dB	      & {\bf38.65dB}       \\
    \hline
\end{tabular}
\end{adjustbox}
\label{tab:EDVRResults}
\end{table}
%============= Tab: Observation 2 =============%

{\flushleft \bf Observation 2:}
\emph{Single frame super-resolution is more suitable for handling patches with temporal redundancy in stationary objects and background.}

% ob2: 单帧超分处理TR更好
Since the temporal redundancy contents is universal in widely-used public videos, we should also investigate whether it will interfere with existing local propagation-based VSR networks.
Following the settings of the experiment in Sec.~\ref{sec:1}, we super-resolve all the Type A and Type B sequences in the validation set with both the EDVR-1f and original EDVR~(EDVR-5f).
The EDVR-1F is modified upon the original EDVR for single frame input, which will be described in Sec.~\ref{sec:4.1}.
As shown in Table~\ref{tab:EDVRResults}, although the EDVR-5f achieves better results on the type B sequences, 
the single frame super-resolution method~(EDVR-1f) can outperform EDVR-5f with lower computational cost on the type A sequences.
% 分析导致实验结果的原因
Since the type A sequences refer to the sequences with temporal redundancy, we analyze that alignment and fusion module of original EDVR may regard these changed pixels due to noisy and information loss during encoding and decoding as the useful temporal information and bring adverse effect on the VSR problem in the sequences with temporal redundancy.
% 得出结论
Therefore, the single frame super-resolution is more suitable for handling patches with temporal redundancy.

%============= Tab: Observation 3 =============%
\begin{table}[!t] 
  \centering
  \caption{\textbf{The performance of BasicVSR in the simulated Type A sequences.} For a fair comparison, the PSNR are calculated on the original dynamic frames.}
  \begin{adjustbox}{width=0.95\linewidth}
  \begin{tabular}{l c c c c c c }
      \hline
      {\bf Training dataset}   & DS  & +10df & +20df & +30df & +40df & +50df     \\
      \hline
       REDS         & 27.38 dB & 27.31 dB  & 27.24 dB & 27.12 dB & 26.99 dB & 26.84 dB \\
       Vimeo        & 25.85 dB & 25.78 dB  & 25.73 dB & 25.69 dB & 25.64 dB & 25.58 dB\\
      \hline  
  \end{tabular}
  \end{adjustbox}
  \label{tab:RNNtest}
  \end{table}
  %============= Tab: Observation 3 =============%

{\flushleft \bf Observation 3:}
\emph{Patches with temporal redundancy in the video sequence will hinder the propagation of non-local propagation-based VSR networks.}

% ob3: TR会阻碍信息的传播
According to the {\bf Observation 2}, the existence of temporal redundancy will bring negative effect to local propagation-based VSR algorithms,
where only local information can be exploited.
On the other hand, non-local propagation-based VSR algorithms can exploit the long-term temporal information by taking all the inference frames as inputs.
To investigate the influence of the temporal redundancy on such longer input sequences,
we conduct an experiment based on the BasicVSR model.
Specifically, we selected 4 downsampling videos with dynamic scenes from the REDS and super-resolve them with the BasicVSR trained on the REDS and the Vimeo respectively.
Then, to simulate the Type A sequence and introduce the temporal redundancy, we randomly choose 10 frames from each video and replicated them several times~(range from 1 to 5), progressively.
For each time, we super-resolve all the extended videos with BasicVSR and record its performance.
As shown in Table~\ref{tab:RNNtest}, the two BasicVSR models both suffer from the performance decline as the length of frames with temporal redundancy increases,
% 分析TR会阻碍信息的传播的原因
which demonstrates limitation of RNN due to the recurrent nature, where one frame may strongly affect the next adjacent frame but its influence is quickly lost after few time steps.
Therefore, despite of long input sequences, the temporal redundancy will still bring negative effect to the RNN-based VSR network by hindering the information propagation.
Similarly results in the realistic video can be found in the supplementary.

\section{Methodology}
\label{sec:4}

% 由ob1得到TR在public video中大量存在, 所以解决TR是必要的
From the observations in Sec.~\ref{sec:3.1}, the temporal redundancy is universal:
almost 70\% patch sequences in the validation set are Type A sequences, which cannot provide any useful information for the VSR algorithms.
Therefore, it's necessary to optimize the existing VSR algorithms to handle the patches with temporal redundancy.
However, there are two categories in the existing VSR methods, which makes it difficult to propose a unified strategy to improve two frameworks simultaneously.
% 由ob2和ob3分别提出Boosted EDVR和Boosted BasicVSR
In this section, based on the {\bf Observation 2} and {\bf Observation 3}, we introduce two effective plug-and-play methods for local and non-local propagation-based networks to optimize the super-resolving process for patches with temporal redundancy.

%============= Fig: Boosted EDVR =============%
\begin{figure*}[!t]
  \centering
  \includegraphics[width=0.9\linewidth]{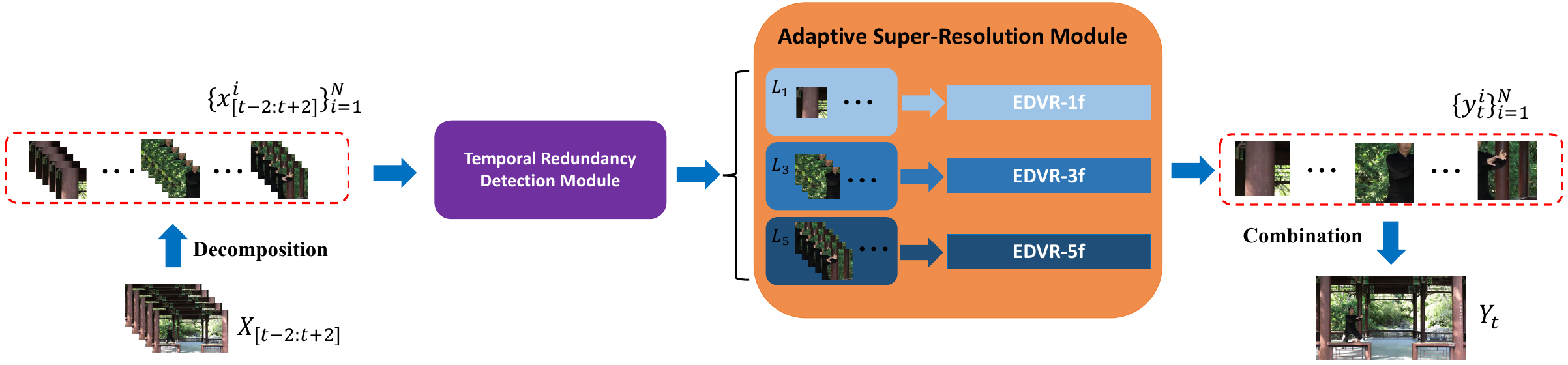}
  \caption{\textbf{Overview of the proposed Boosted EDVR.} 
  }
\label{fig:EDVR}
\end{figure*}
%============= Fig: Boosted EDVR =============%

\subsection{Boosting Local Propagation-Based Networks}
\label{sec:4.1}

% 由ob2提出Boosted EDVR
The local propagation-based VSR methods~\cite{wang2019edvr,SPMC,DUF,VESPCN} take LR images within a local window as inputs and employ the local information for restoration. 
However, based on the \textbf{Observation 2}, the patches with temporal redundancy should be specially handled.
To achieve this, we try to introduce a temporal redundancy detection module to the existing methods and super-resolve each patch adaptively. 
In the following parts, we will use the EDVR as example to show how the proposed plug-and-play method optimize the local propagation-based VSR methods.

% Boosted EDVR的总体描述
Inspired by the recent work, Class-SR\cite{classsr}, we extend the original EDVR to a new pipeline, namely \textbf{Boosted EDVR}, to perform temporal redundancy detection and super-resolution simultaneously.
As shown in Figure~\ref{fig:EDVR}, the proposed \textbf{Boosted EDVR} consists of two modules: Temporal Redundancy Detection Module~(TRDM) and Adaptive Super-Resolution Module~(ASRM).
The input five LR neighboring frames $X_{[t-2:t+2]}$ are first decomposed into $N$ overlapping patch sequences $\{x^{i}_{[t-2:t+2]}\}^{N}_{i=1}$.
Then, each decomposed patch sequence $x^{i}_{[t-2:t+2]}$ is fed to the TRDM and assigned a movement label~($L^{i}_{j}, j\in\{1,3,5\}$) according to its motion state among neighboring patches.  
After that, all the patch sets with the same label will be concatenated in the batch-size dimension and super-resolved by the optimal EDVR model in ASRM.
Finally, we combine all the super-resolved patches $\{y^{i}_{t}\}^{N}_{i=1}$ to get the final SR results $Y_{t}$.

\noindent{\bf Temporal redundancy detection module.}
The goal of TRDM is to detect the temporal redundancy and assign a movement label to each patch sequence.
Based on the {\bf Observation 2}, the temporal redundancy exists in the stationary objects and background, which means we should find a way to represent the motion state between two patches.
Since the optical flow is a widely-used metric to describe the motion information, we use the mean values of the optical flow to represent the motion state, which can be formulated as:
\begin{equation}\label{eq:1} 
  m^{i}_{-1\rightarrow 0} =  mean(\vert f(x^{i}_{t-1}, x^{i}_{t})\vert),
\end{equation}
where $f$ denotes the optical flow estimator, $\vert \cdot \vert$ denotes absolute value, $mean$ is the mean value, and $m^{i}_{-1\rightarrow 0}$ denotes the motion state between the reference patch~($x^{i}_{t}$) and its neighboring patch ($x^{i}_{t-1}$) in the patch sequence $i$. 
We choose the traditional DIS~\cite{DIS} algorithm as the optical flow estimator since it only slightly increase the computational cost.

% Temporal redundancy detection module.
For $i$-th patch sequence, we successively calculate the motion states of all the neighboring patches, which denote as $m^{i}_{-2\rightarrow -1}$, $m^{i}_{-1\rightarrow 0}$, $m^{i}_{1\rightarrow 0}$, and $m^{i}_{2\rightarrow 1}$).
Then, we assigned a movement label~($L_{j}(x^{i}_{[t-2:t+2]}), j\in\{1,3,5\}$) according to these motion states:
\begin{align}
  L^{i}_{j} = \left\{\begin{aligned}
  L^{i}_{1}~~~~  &\text{if}~m^{i}_{-1\rightarrow 0}<\gamma~~\text{and}~~m^{i}_{1\rightarrow 0}< \gamma,
  \\
  L^{i}_{3}~~~~  &\text{elif}~m^{i}_{-2\rightarrow -1}<\gamma~~\text{and}~~m^{i}_{2\rightarrow 1}< \gamma,
  \\
  L^{i}_{5}~~~~  &\text{otherwise,}
\end{aligned}\right.\end{align}
where $\gamma$ is the threshold to discriminate the patch with stationary objects and background and $L^{i}_{j}$ denotes $j$ dynamic patches involved in $i$-th patch sequence.
With TRDM, we can determine which model in the following ASRM should be used to obtain better super-resolved results.

\noindent{\bf Adaptive Super-Resolution Module.}
The ASRM, which consists of the original EDVR~(EDVR-5f) and two of its variants~(EDVR-3f and EDVR-1f), is designed to super-resolve each patch sequence with the optimal model.
Specifically, we adopt the EDVR-1f model, which is modified for single frame input based on EDVR, to super-resolve all the patch sets with the movement label $L_{1}$, 
since there is no useful temporal information in the neighboring patches.
Similarly, the EDVR-3f model and EDVR-5f model will process the patch sequences with the movement labels $L_{3}$ and $L_{5}$, respectively.
Different from experiment in {\bf Observation 2}, we introduce the EDVR-3f model by taking the situation that the temporal redundancy only occurs at the border frames of the patch sequence into consideration.

% Adaptive Super-Resolution Module.
To acquire EDVR-1f and EDVR-3f with minimal modification, we only slightly changes the forward flow of the original EDVR~(EDVR-5f) without any changes on the network architecture.
For EDVR-1f and EDVR-3f, the PCD alignment module and the temporal attention layers in TSA module are only performed once and threes times, respectively, 
and the features will be replicated to the same shape as EDVR-5f before sending to the fusion convolutional layer in the TSA module.
Since we remove the unnecessary calculation in the PCD alignment and TSA modules of the EDVR-1f and EDVR-3f, the proposed pipeline will be more efficient than the original EDVR. 
More detailed of the EDVR-1f and EDVR-3f can be found in the supplementary.
To ensure the EDVR-1f and EDVR-3f can achieve comparable super-resolving ability as EDVR, we also fine-tune them on the same training dataset~(REDS) and with same hyper-parameter as EDVR. 
Experiments show that such a simple pipeline can improve the performance of EDVR close to the upper bound with less FLops.

%============= Fig: Boosted BasicVSR =============%
\begin{figure*}[!t]
  \centering
  \includegraphics[width=0.7\linewidth]{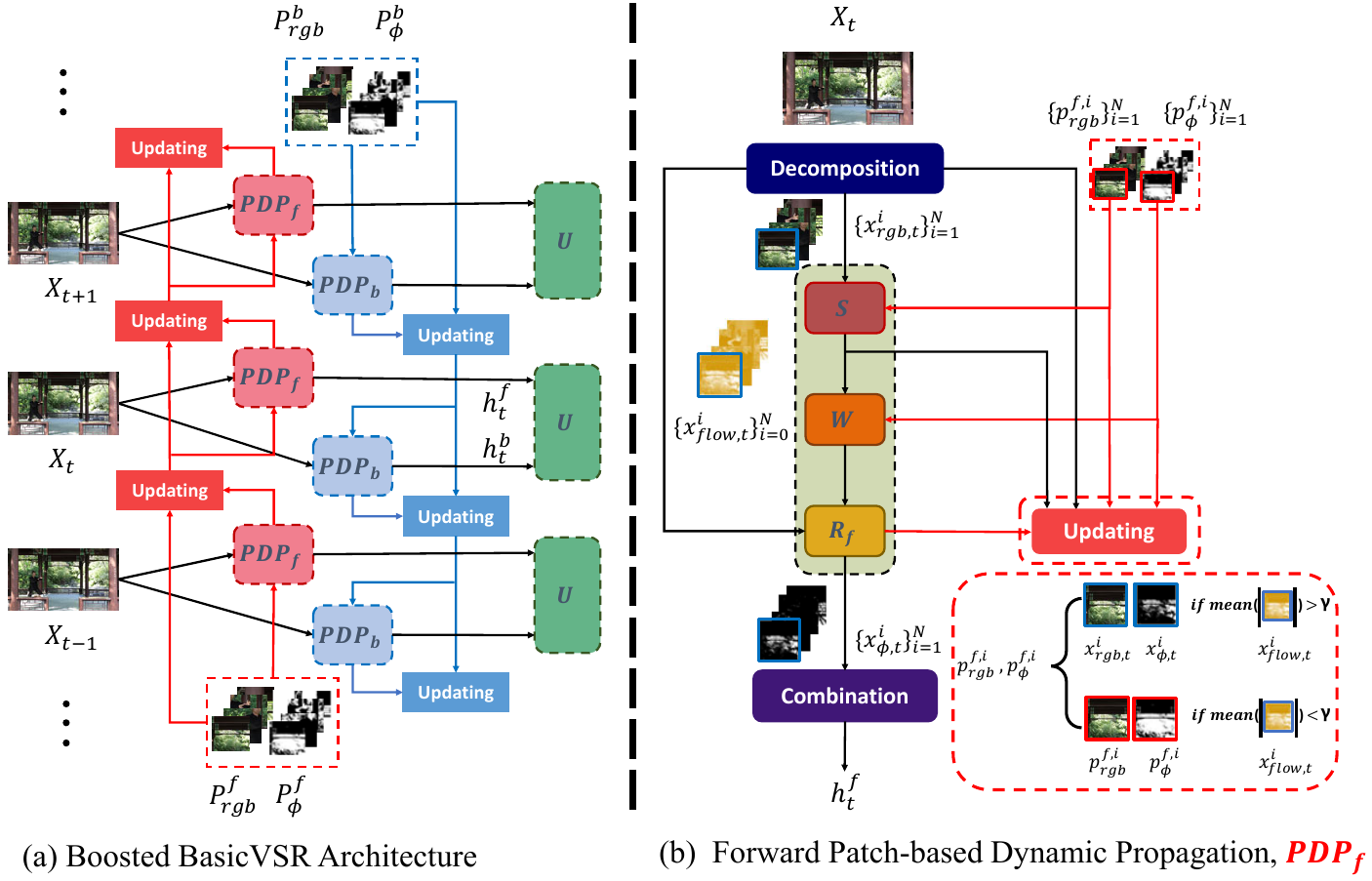}
  \caption{\textbf{Overview of the proposed Boosted BasicVSR.} 
  (a) The upsampling module $U$ contains multiple pixel-shuffle and convolutions. 
  The \textbf{red} and \textbf{blue} colors represent the forward and backward propagations, respectively. 
  (b) $S$, $W$, and $R_{f}$ refer to the flow estimator, spatial warping module, and residual blocks for forward branch, respectively.
  }
\label{fig:BasicVSR}
\end{figure*}
%============= Fig: Boosted BasicVSR =============%   

\subsection{Boosting Non-Local Propagation-Based Networks}
\label{sec:4.2}

% 由ob3提出Boosted BasicVSR
Unlike local propagation-based methods, the non-local propagation-based methods can exploit long-term information by taking all the inference frames as inputs and sequentially propagation.
However, based on the \textbf{Observation 3}, the patches with temporal redundancy in the video sequence will hinder the propagation, which inevitably limits the potential of the existing non-local propagation-based VSR methods.
To better exploit the long-term information, we propose a new plug-and-play method by introducing a Patch-based Dynamic Propagation~(PDP) branch to dynamically propagate the long-term information in a patch-wise way.
As shown in Figure~\ref{fig:BasicVSR}(a), we deploy the proposed plug-and-play method to BasicVSR, namely \textbf{Boosted BasicVSR}, by replacing the original propagation branches with the proposed PDP branches.
In the following parts, we will show how the PDP branch works in forward propagation~($PDP_{f}$), and the PDP branch in the backward propagation~($PDP_{b}$) can be derived accordingly.

% PDP的总体描述
Unlike the propagation branch in the BasicVSR, the proposed forward PDP branch adopts dynamical propagation, where each patch of the current frame can receive information from different frames. 
To achieve this, the proposed forward PDP branch maintains a patch pool $P^{f}_{rgb}$ and its corresponding hidden state pool $P^{f}_{\phi}$ to restore the useful information of patches from different frames.
Then, the forward PDP branch takes the current LR frame $X_{t}$, $P^{f}_{rgb}$, and $P^{f}_{\phi}$ as inputs and generates the forward features $h^{f}_{t}$ while updating $P^{f}_{rgb}$ and $P^{f}_{\phi}$ based on the temporal redundancy detection.
The advantage of maintaining an independent patch-wise hidden states pool and propagating it to current frame instead of the neighboring hidden states is that the useful information in the long-term frame can directly connect to current frame without accumulating useless redundancy information.
The detail of the PDP branch is shown in Figure~\ref{fig:BasicVSR}(b), which consists of two stages: features aggregation and patch pools update.

\noindent{\bf Features aggregation.}
This stage is design to aggregate the information in the maintained pools~($P^{f}_{rgb}$ and $P^{f}_{\phi}$) with the current frame.
The $P_{\phi}$ and $P_{rgb}$ contains $N$ overlapping patches $\{p^{f,i}_{rgb}\}^{N}_{i=1}$ and corresponding hidden state patches $\{p^{f,i}_{\phi}\}^{N}_{i=1}$ which are sorted by their positions in the frame.
Since the BasicVSR adopts the first-order propagation, we only maintain the information of one patch for each position.

% Features aggregation.
To estimate the optical flow for spatial alignment of the hidden state pool $P^{f}_{\phi}$, we first decompose current frame $X_{t}$ into $N$ overlapping patches~($\{x^{i}_{rgb,t}\}^{N}_{i=1}$).
Then, the optical flows of all the patches~($\{x^{i}_{flow,t}\}^{N}_{i=1}$) are calculated by sending the correspond patches in the $\{x^{i}_{rgb,t}\}^{N}_{i=1}$ and $\{p^{f,i}_{rgb}\}^{N}_{i=1}$ to the optical estimator~($S$).
After that, we perform warping~($W$) on the patches in the hidden state pool using the estimated flow for the further refinement in the residual blocks $R_{f}$.
By feeding the warped hidden state pool and the overlapping patches of current frame into the residual blocks, the intermediate features patches of current frame~($\{x^{i}_{\phi,t}\}^{N}_{i=1}$) can be obtained.
Finally, the forward features $h^{f}_{t}$ can be obtained by combining the $\{x^{i}_{\phi,t}\}^{N}_{i=1}$.

\noindent{\bf Patch pools update.}
In this stage, we try to update the patch pool $P^{f}_{rgb}$ and hidden state pool $P^{f}_{\phi}$ with the information of current frame.
As shown in Figure~\ref{fig:BasicVSR}(b), we use a similar temporal redundancy detection method in \textbf{Boosted EDVR} to decide which patches in $P^{f}_{rgb}$ and $P^{f}_{\phi}$ should be updated with current frame.
Since we already obtain the optical flows ($\{x^{i}_{flow,t}\}^{N}_{i=1}$) in the features aggregation stage, we direct use Equ.~\eqref{eq:1} to obtain the motion states $\{m^{i}_{t}\}^{N}_{i=1}$~(\textbf{the box with red dashed line}) of all the corresponding patches between $\{p^{f,i}_{rgb}\}^{N}_{i=1}$ and $\{x^{i}_{rgb,t}\}^{N}_{i=1}$.
Then, to ensure the useful information can be accumulated, each patch set~($p^{f,i}_{rgb}$ and $p^{f,i}_{\phi}$) in the two pools will be replaced by the information of corresponding patch of current frame~($x^{i}_{rgb,t}$ and $x^{i}_{\phi,t}$) when the motion state of this patch~($m^{i}_{t}$) is larger than the threshold ${\gamma}$.
Otherwise, which means temporal redundancy exists in these two patch, the information of current frame will be discarded to avoid vanishing the useful information.
Finally, the updated pools will be propagated to the next frame.
Experiments demonstrate that the proposed PDP scheme can significantly improve the performance of BasicVSR and solve the generalization ability without any training process.

\section{Experiments}
\label{sec:5}
\subsection{Datasets and Settings}
\label{sec:5.1}

% 实验选择的数据集
Since we aim to boost existing VSR algorithms with minimal modifications, we only fine-tune the EDVR-1f and EDVR-3f on the training set of the REDS.
Then, we use the REDS4, Vid4, and the DTVIT as the test set to compare the proposed models with existing VSR algorithms.
For fair comparison, all the evaluated models are trained and tested on the dataset with 4$\times$ bicubic downsampling.

% 实验设置
For the proposed two methods, each LR frame is decomposed into 64$\times$64 patches with stride 56 (with 8 pixel overlaps), and the combination operation combines all the patches to an integrated frame by averaging overlapping areas.
The threshold $\gamma$ in \textbf{Boosted EDVR} and \textbf{Boosted BasicVSR} are set to 1 and 0.2, respectively. 
The DTVIT dataset and source code will be made available to the public.

\subsection{Experiments on Network Configurations}
\label{sec:5.2}

% 消融实验的介绍
To investigate the effect of different network configurations and find the optimal one for the proposed \textbf{Boosted EDVR} and \textbf{Boosted BasicVSR}, we evaluate several models with alternative configurations. When evaluating \textbf{Boosted EDVR}, we also calculate the average FLOPs to evaluate the efficiency. For quick verification during the design stage, we still select the validation set of the DTVIT dataset for the ablation study.

%%%%%%%%%%%%%%%%%%%%%%%%%%%%%%%%%%%%%%%  Ablation Study 1  %%%%%%%%%%%%%%%%%%%%%%%%%%%%%%%%%%%%%%%
\begin{table}[!t]
\large
  \centering
  \caption{\textbf{Analysis on each component of the proposed Boosted EDVR.}}
  \begin{adjustbox}{width=1.0\linewidth}
    \scalebox{1}{
    \begin{tabular}{rcccccc}
      \hline
      \textbf{Methods}        & EDVR        & TR-EDVR    & Boosted EDVR-(15)    & Boosted EDVR-(135)    & Boosted EDVR-(UB)        \\
      \hline
      EDVR-5f                 & \checkmark      &  \checkmark     &  \checkmark          &  \checkmark           &  \checkmark              \\
      EDVR-3f                 &               &                 &                      &  \checkmark           &  \checkmark                \\
      EDVR-1f                 &               &  \checkmark     &  \checkmark          &  \checkmark           &  \checkmark             \\
      TR detection            &              &  \checkmark     &  \checkmark          &  \checkmark           &                      \\
      DIS flow                &             &                 &  \checkmark          &  \checkmark           &                  \\
      \hline
      Flops                   & 758M~(100\%)   & 661M~(87\%)    & 522M~(69\%)       &  \textbf{519M~(68\%)} & 622M~(82\%)                        \\
      PSNR                    & 33.42       & 34.30           & 34.50            & \textbf{34.51}  & 34.72                    \\                 
      \hline
    \end{tabular}}
  \end{adjustbox}
  \label{tab:ablation1}
\end{table}
 %%%%%%%%%%%%%%%%%%%%%%%%%%%%%%%%%%%%%%%  Ablation Study 1 %%%%%%%%%%%%%%%%%%%%%%%%%%%%%%%%%%%%%%%

\noindent{\bf Study of the \textbf{Boosted EDVR}.}
Starting from the original EDVR, we first use the mean square errors~(MSE) of pixel values to detect temporal redundancy
and use the fine-tuned EDVR-1f models to super-resolve the patch sequences with movement label $L_{1}$.
We denote these configuration as TR-EDVR.
As shown in Table~\ref{tab:ablation1}, the TR-EDVR can achieve 0.88 dB performance gain over the original EDVR with less FLOPS.
These results also demonstrate the effectiveness of the proposed pipeline with temporal redundancy detection module, which can adaptively super-resolve different patch sets with the optimal model.
Since the optical flow is widely used to describe the motion information, we use the mean values of the DIS optical flow to represent the motion state and form the Boosted EDVR-(15).  
The Boosted EDVR-(15) outperforms TR-EDVR by a margin of 0.2 dB while the overall FLOPs drop from 661M to 522M, which demonstrates that the DIS optical flow is more suitable for redundancy detection than MSE.
Futhermore, we also introduce a fine-tuned EDVR-3f model, namely Boosted EDVR-(135), to super-resolve the patch sets where the temporal redundancy only occurs at the border patches~($m^{i}_{-2\rightarrow -1}<\gamma~~\text{and}~~m^{i}_{2\rightarrow 1}<\gamma$).
Since both the performance and efficiency are slight improved by introducing EDVR-3f, we choose the Boosted EDVR-(135) as the final configurations of \textbf{Boosted EDVR}.
Compared with the EDVR, the proposed \textbf{Boosted EDVR} can achieve 1.09 dB performance gain with only 68\% computational cost.
Finally, we also obtain the upper bound of the \textbf{Boosted EDVR} by simultaneously feeding each patch sequence to three models respectively and choosing the best one~(in terms of PSNR) as result.
Since the performance gap between the \textbf{Boosted EDVR} and the upper bound model is relatively small~(0.21 dB) and the proposed method can save more computational cost, we think our pipeline is acceptable by maintaining a good balance between the performance and efficiency. 

%%%%%%%%%%%%%%%%%%%%%%%%%%%%%%%%%%%%%%%  Ablation Study 2  %%%%%%%%%%%%%%%%%%%%%%%%%%%%%%%%%%%%%%%
\begin{table}[!t]
    \large
      \centering
      \caption{\textbf{Analysis on each key factor of the proposed PDP branch.}}
      \begin{adjustbox}{width=0.85\linewidth}
        \scalebox{1}{
        \begin{tabular}{rcccc}
          \hline
          \textbf{Methods}   & BasicVSR  & TR-BasicVSR    & DP-BasicVSR  & Boosted BasicVSR       \\
          \hline
          Frame type classification &   & \checkmark  &  \checkmark     &  \checkmark \\
           Dynamic propagation   &   &             &  \checkmark     &  \checkmark       \\
          Patch-wise            &   &  &  &  \checkmark       \\
          \hline
          PSNR                 & 27.96   & 32.57  & 33.22 & {\textbf{34.08}}             \\                
          \hline
        \end{tabular}}
      \end{adjustbox}
      \label{tab:AB-2}
  \end{table}
%%%%%%%%%%%%%%%%%%%%%%%%%%%%%%%%%%%%%%%  Ablation Study 2  %%%%%%%%%%%%%%%%%%%%%%%%%%%%%%%%%%%%%%%  

\noindent{\bf Study of the \textbf{Boosted BasicVSR}.}
In this part, we will evaluate the importance of three key factors in the proposed Patch-based Dynamic Propagation~(PDP): temporal redundancy detection, dynamic propagation, and patch-wise strategy.
As shown in Table~\ref{tab:AB-2}, the performance of BasicVSR trained on the REDS is much worse than the original EDVR~(27.96 dB vs. 33.42 dB) on the validation set, which is contradictory to the results on the existing datasets.
We owe this severe generalization problem of BasicVSR to the error accumulation of the optical flow: 
since the optical flow estimator in BasicVSR may regard these changed pixels due to noisy and information loss during encoding and decoding as the useful temporal information, it will produce inaccurate optical flow between the frames with stationary objects and background and the error will be accumulated through the propagation~(\textbf{more analysis can be found in the supplementary}).
To overcome this problem, we propose a new pipeline, namely TR-BasicVSR, to super-resolve stationary and dynamic frames separately.
More specifically, we follow the sequence definitions in {\bf Observation 1} and divide the types of each test video using the redundancy detection module in Sec.~\ref{sec:4.1}.
Then, we combine all the Type B sequences into one sequence and super-resolve it with BasicVSR.
For Type A sequences, where all the frames are similar in one sequence, we super-resolve each frame independently to avoid the error accumulation of the optical flow.
As shown in Table~\ref{tab:AB-2}, the TR-BasicVSR obtain significant performance gain over the original BasicVSR, which demonstrates that the temporal redundancy detection can solve the generalization problem effectively.

% TR-BasicVSR到DP-BasicVSR
However, the TR-BasicVSR cannot exploit any temporal information from the Type B sequences when handling Type A sequences, which inevitably limits its performance.
Therefore, we further introduce the dynamic propagation scheme to TR-BasicVSR~(referred to as the DP-BasicVSR) and make sure each frame can exploit the useful temporal information.
Specifically, the DP-BasicVSR maintains an anchor frame and its corresponding hidden states to restore the long-term information from the closest dynamic frame and propagate it to current frame.
Since the dynamic propagation scheme can directly propagate the information from the long-term frame to current frame without accumulating useless redundancy information of the stationary objects and background, 
the DP-BasicVSR outperforms TR-BasicVSR by a margin of 0.65 dB.

% DP-BasicVSR到Boosted BasicVSR
Finally, due to the contents in different patches of a video may changes independently,
the final \textbf{Boosted BasicVSR} maintain a patch pool $P^{f}_{rgb}$ and its corresponding hidden state pool $P^{f}_{\phi}$ to restore long-term information in a patch-wise way.
By adopting the patch-wise strategy, the \textbf{Boosted BasicVSR} achieves 0.86 dB performance gain over DP-BasicVSR.
Overall, the proposed \textbf{Boosted BasicVSR} can solve the generalization problem of the pre-trained BasicVSR and boost its performance without any training process, which demonstrates the effectiveness of the proposed PDP scheme.

%%%%%%%%%%%%%%%%%%%%%%%%%%%%%%%%%%%%%%%  Comparison Study  %%%%%%%%%%%%%%%%%%%%%%%%%%%%%%%%%%%%%%% 
\begin{table}[!t] 
  \centering
  \caption{{\bf Quantitative comparison (PSNR/SSIM).}
  All results are calculated on RGB-channel.}
  \begin{adjustbox}{width=0.85\linewidth}
  \begin{tabular}{l c c c c}
      \hline
      \multirow{2}{*}{\bf Training dataset} & \multirow{2}{*}{\bf Methods}   & REDS Val          & Vid4            & DTVIT        \\
      &   & PSNR/SSIM         & PSNR/SSIM       & PSNR/SSIM    \\
      \hline
      \multirow{9}{*}{REDS} &
      Bicubic           & 26.14/0.7292                     & 23.78/0.6347                      & 29.46/0.8870   \\ &
      DUF              & 28.63/0.8251                     & 18.45/0.5117                      & 23.17/0.6517   \\ &
      RBPN              & 30.09/0.8590                     & 25.66/0.8029                      & 32.74/0.9208   \\ &
      MuCAN             & 30.88/0.8750                      & 25.33/0.7994                      & 30.58/0.9072   \\ &
      EDVR            & 30.53/0.8699                     & 25.34/0.7951                      & 32.00/0.9205   \\ &
      EDVR-L            & 31.09/0.8800                     & 25.40/0.8008                      & 32.39/0.9277  \\ &
      BasicVSR          & 31.42/0.8909 & 25.75/0.8155    & 27.13/0.8165   \\ &
      \textbf{Boosted EDVR}    & 30.53/0.8699                     & 25.32/0.7950                      & 32.91/0.9262   \\ &
      \textbf{Boosted BasicVSR}  & {\textbf{31.42/0.8917}}   & {\textbf{25.93/0.8202}}      & {\textbf{33.21/0.9340}}    \\
      \hline
      \multirow{2}{*}{Vimeo} &
      BasicVSR   & 30.32/0.8672 & 25.82/0.8085 & 33.31/0.9368   \\ &
      \textbf{Boosted BasicVSR}  & {\textbf{30.32/0.8673}} & {\textbf{25.84/0.8093}} & {\textbf{33.79/0.9503}}    \\
      \hline
  \end{tabular}  
  \end{adjustbox}
  \label{tab:comparison}
  \end{table}
%%%%%%%%%%%%%%%%%%%%%%%%%%%%%%%%%%%%%%%  Comparison Study  %%%%%%%%%%%%%%%%%%%%%%%%%%%%%%%%%%%%%%% 

\subsection{Comparisons with Existing VSR algorithms}
\label{sec:5.3}

% 对比实验所选择的方法
To further evaluate the proposed methods,
we conduct comprehensive experiments by comparing \textbf{Boosted EDVR} and \textbf{Boosted BasicVSR} with several state-of-the-art VSR algorithms: 
DUF, RBPN, MuCAN, EDVR, EDVR-L, and BasicVSR.

% REDS的相关对比实验，说明Boosted方法的鲁棒性
The first and second columns in Table~\ref{tab:comparison} show the quantitative results on the REDS and Vid4, where all the testing videos are first-person videos with consistent movement.
As expected, the proposed \textbf{Boosted EDVR} and \textbf{Boosted BasicVSR} only achieve comparable performance with EDVR and BasicVSR on these two datasets, since they are optimized for videos with temporal redundancy.
However, the stable performance on the first person videos demonstrates that the proposed methods are robustness and will not bring any adverse influence to existing datasets.

% 使用更大的REDS测试集来做Boosted BasicVSR和BasicVSR的对比实验
Meanwhile, for lager evaluation dataset, we select six clips in the REDS training clips and extend the REDS test set~(i.e. REDS4) to ten clips, denoted by REDS10. The remaining training clips are used as new training dataset~(a total of 260 clips). Based on the setting above, the proposed \textbf{Boosted BasicVSR} trained on the REDS outperforms BasicVSR by a margin of 0.15 dB on the REDS10~(30.86dB v.s. 31.01dB), which demonstrate that the proposed \textbf{Boosted BasicVSR} can still improve the performance on the REDS and temporal redundancy also exists in some of the REDS.

% DTVIT的相关实验，说明动态数据集和静态数据集中存在很大的泛化问题，能用所提方法解决
To comprehensively evaluate the performance of VSR algorithms on different types of public videos, we also evaluate these algorithms on the collected DTVIT dataset.
As shown in the third column of Table~\ref{tab:comparison}, the BasicVSR trained on the REDS performs not well on the collected video dataset due to the generalization problem.
Although the EDVR-M and EDVR-L achieve favorable performance than other methods, 
the proposed \textbf{Boosted EDVR} can further improve the performance by up to 0.91 dB over EDVR-M and outperform EDVR-L with much lower computational cost.
Moreover, the proposed \textbf{Boosted BasicVSR} can solve the generalization problem and significantly improve the performance by a large margin of 6.28 dB over BasicVSR. 
In addition, the \textbf{Boosted BasicVSR} outperforms \textbf{Boosted EDVR} by a margin of 0.24 dB with comparable computational cost, which coincides with the results on the existing dataset. 
Overall, both \textbf{Boosted EDVR} and \textbf{Boosted BasicVSR} are able to achieve remarkable performance on the collected dataset, 
which demonstrates that the proposed plug-and-play methods can improve the performance and robustness of existing VSR algorithms.

% 排除泛化问题(Vimeo模型的结果)后的提升
To verify the proposed method can not only solve the generalization problem but also can enhance the effectiveness of the propagation branches, we also apply the proposed Patch-based Dynamic Propagation branch to the BasicVSR trained on the Vimeo to see whether the improvement can be obtained.
As shown in Table~\ref{tab:comparison}, since the Vimeo contains more types of videos, the BasicVSR trained on it will not suffer of severe generalization problem and achieves favorable on the DTVIT dataset.
In addition, the proposed \textbf{Boosted BasicVSR} can still outperform the BasicVSR without any training process, which demonstrates that the proposed method can effectively improve the performance of existing non-local propagation-based VSR algorithms.
Moreover, our method largely boosts the BasicVSR trained on the REDS and achieves similar performance with the BasicVSR trained on the Vimeo~(33.15 dB v.s. 33.31 dB) without extra training datasets and time-consuming training process, which makes our method practical in real-world applications and can be easily extended to other video restoration tasks.

%============= Fig: Qualitative comparison =============%
\begin{figure}[!t]
  \centering
  \includegraphics[width=0.95\linewidth]{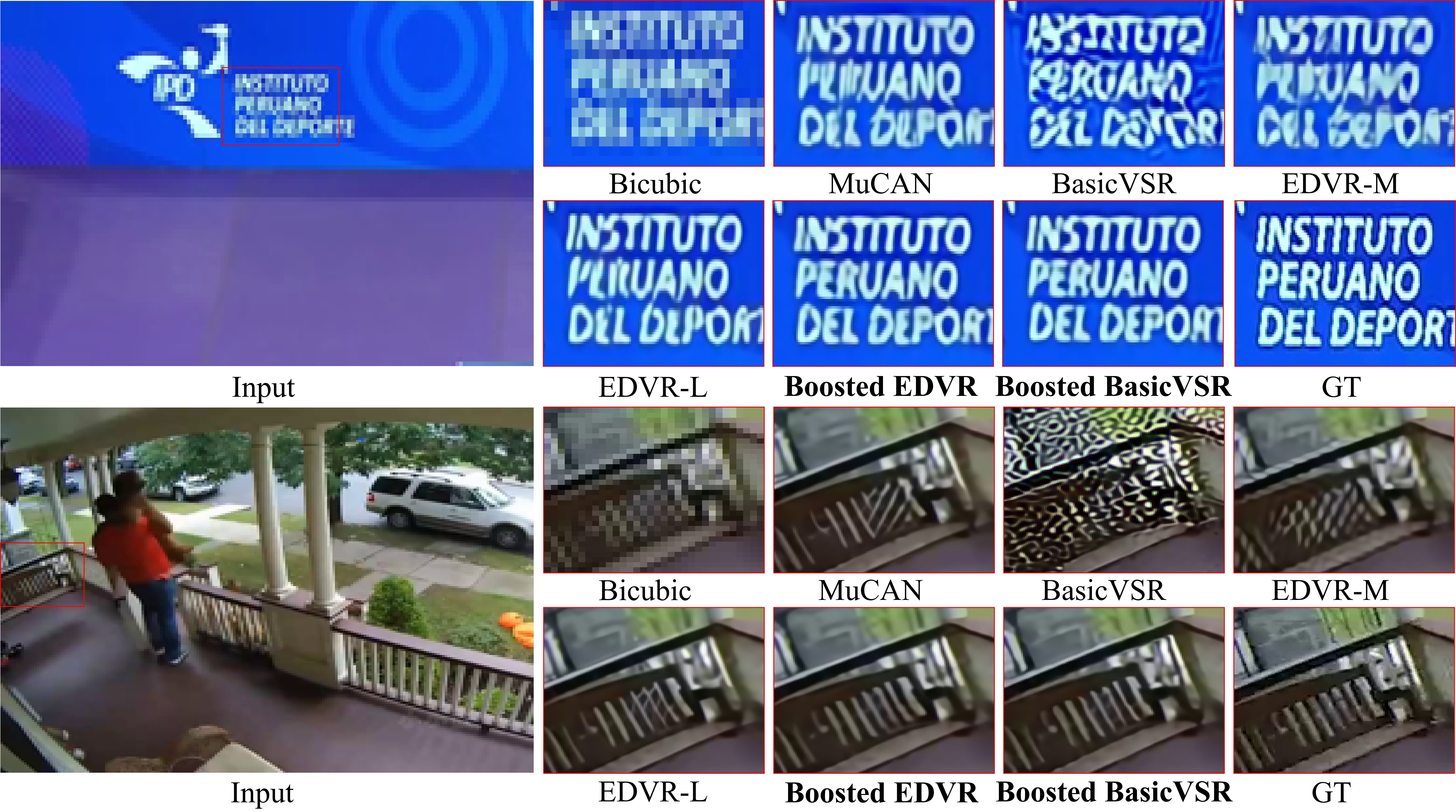}
  \caption{\textbf{Qualitative comparison on the DTVIT dataset.}}
\label{fig:Qualitative comparison}
\end{figure}
%============= Fig: Qualitative comparison =============%

% 定性比较(结果对比图)
Qualitative comparisons are shown in Figure~\ref{fig:Qualitative comparison}. 
The \textbf{Boosted EDVR} and \textbf{Boosted BasicVSR} recover finer details and sharper texts in the videos from the DTVIT dataset. 
More examples are provided in the supplementary.

%-------------------------------------------------------------------------
\section{Conclusion}
\label{sec:6}

% TR的重要性
In this paper, we investigate the temporal redundancy in the video and note it as an important factor for VSR methods for three reasons:
(1) it will bring unnecessary computational cost for local propagation-based networks (e.g., EDVR), 
(2) it will cause severe generalization problem for the models trained on the dynamic datasets (as BasicVSR trained on the REDS performs not well in the DTVIT), 
and (3) it will gradually vanish the useful temporal information in the distant frame and hinder the performance of the non-local propagation-based networks.

% 提出方法总结
Therefore, we focus on optimizing the existing VSR algorithms by taking the adverse effect of the temporal redundancy into consideration.
Through introducing a temporal redundancy detection and adaptive super-resolution module to the original EDVR, we propose \textbf{Boosted EDVR},
a simple yet effective method can improve the performance and accelerate the inference time simultaneously. 
We also propose \textbf{Boosted BasicVSR} by adopting a Patch-based Dynamic Propagation~(PDP) scheme to solve the generalization problem of the original BasicVSR and boost its performance without any training process.
Extensive evaluations show that the proposed modifications can largely improve the performance on the collected dataset without any adverse influence to existing datasets.
We believe that these two plug-and-play methods can also be applied to others video restoration tasks since the temporal redundancy is universal in most public videos.

\bibliography{aaai23}

\end{document}

\maketitle

\centerline{\large{\textbf{Overview}}}
In this supplemental material, we provide more details about the proposed DTVIT dataset in Sec.~\ref{sec:1}.
%Similarly results can be also observed in the realistic video, which can be found in the supplementary.
To further validate the \textbf{Observation 3}, we conduct an experiment on a realistic public video and find similarly observation in Sec.~\ref{sec:2}.
The details of EDVR-1f and EDVR-3f and their differences with EDVR-5f are provided in Sec.~\ref{sec:3}.
We also analysis the generalization problem of the pre-trained BasicVSR on the DTVIT dataset in Sec.~\ref{sec:6}, 
the inaccurate estimated optical flow cause accumulated errors which leads to performance decline.
To better illustrate the efficiency of the proposed method, the evaluations of computational cost and inference time are provided in Sec.~\ref{sec:5}.
Moreover, to verify the universality of our methods, we also deploy the proposed plug-and-play methods to the models trained on the DTVIT-Train~(a newly collected training dataset according to the categories of the DTVIT dataset) in Sec.~\ref{sec:4}.
Finally, more qualitative results on the DTVIT dataset and real-world videos are provided in Sec.~\ref{sec:7}.

\section{More Details about DTVIT}
\label{sec:1}

As mentioned in the manuscript, we collect 96 videos with high-quality and high-resolution as ground-truth and obtain the corresponding low-resolution inputs with 4$\times$ bicubic downsampling.

\begin{figure}[!t] 
  \centering
  \includegraphics[width=0.95\linewidth]{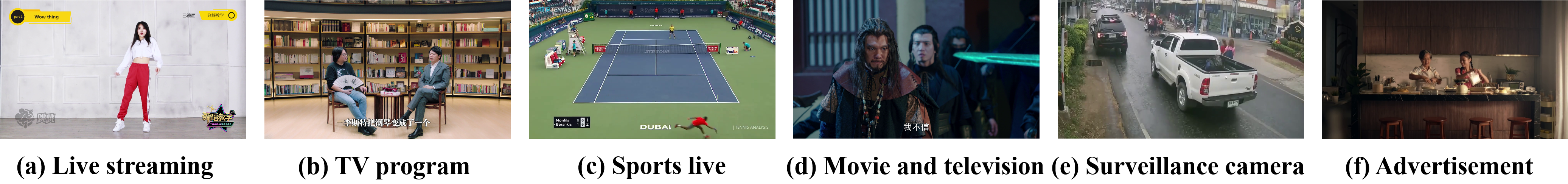}
  \caption{\textbf{Example of the collected DTVIT dataset.} Best viewed on a high-resolution display.}
\label{fig:exampleFigures}
\end{figure}

The types of the videos in the DTVIT dataset are diverse and can be grouped into 7 categories: live streaming, TV program, sports live, movie and television play, surveillance video, advertisement, and first-person video with irregular trajectories. 
% DTVIT数据集的例图
The examples of the collected DTVIT dataset are shown in Fig.~\ref{fig:exampleFigures}
and the specific number of each type is shown in Table~\ref{tab:testset}.

Moreover, we also collect a new training dataset according to the categories of the DTVIT dataset, referred as to DTVIT-Train, to verify the proposed method can not only solve the generalization problem but also can enhance the effectiveness of the propagation branches.
The DTVIT-Train contains 100 video clips and each video clip is consist of 100 frames.

% DTVIT数据集的例图

% 统计表
\begin{table}[!h]
\centering
\caption{{\bf The number of videos in the DTVIT dataset.} Each video contains 100 frames.}
\begin{adjustbox}{width=0.7\linewidth}
\begin{tabular}{c c}
    \hline
    {\bf Video Style}    & Number of videos         \\
    \hline
    Live streaming                         & 19            \\
    TV program                             & 20              \\
    Sports live                            & 18                \\
    Movie and television play              & 9               \\
    Surveillance video                     & 10                \\
    Advertisement                          & 8               \\
    First person video                     & 12              \\
    \hline
    Total                                  & 96                \\
    \hline
\end{tabular}
\end{adjustbox}
\label{tab:testset}
\end{table}

\begin{figure*}[!h]  
  \centering
  \includegraphics[width=0.7\linewidth]{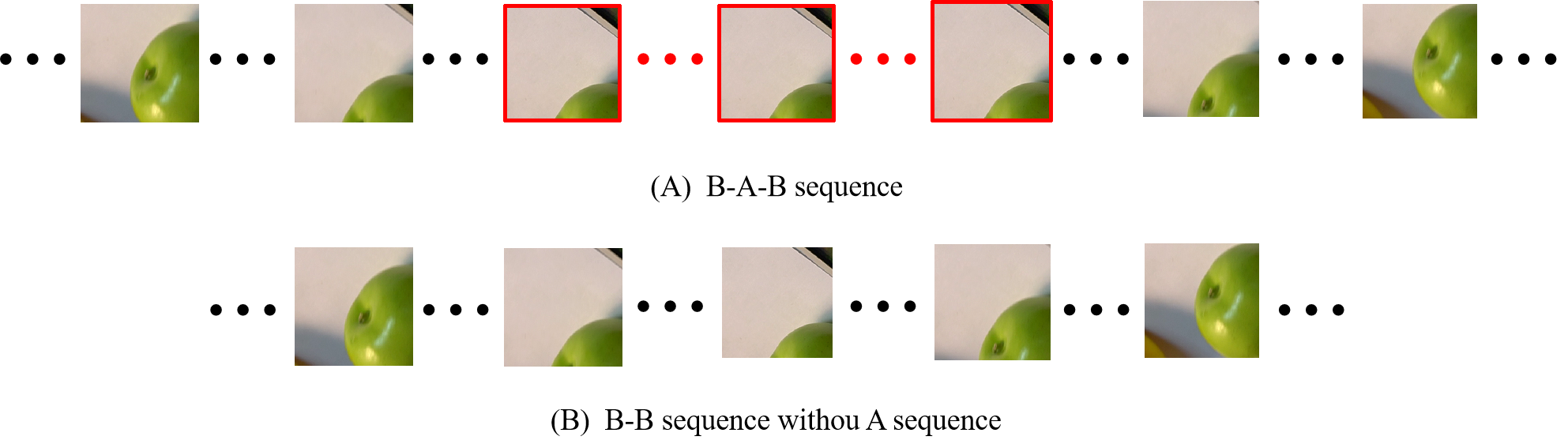}
  \caption{\textbf{Examples of the B-A-B and B-B sequences.}}
\label{fig:BAB&BB}
\end{figure*}

\section{More verification on Observation 3}
\label{sec:2}

In the manuscript, the \textbf{Observation 3} is validated in a simulated Type A sequence, where the temporal redundancy in a long sequence will still hinder the propagation of non-local propagation-based VSR networks.
To further validate this, we apply the BaiscVSR~\cite{basicvsr} on ten patch sequences from the realistic public video in the DTVIT dataset.
%
% We owe this performance declines that the existence of temporal redundancy in input sequences may blocks the propagation of useful information.
%
As shown in Fig.~\ref{fig:BAB&BB}-(a), all the selected patch sequences have a B-A-B type, which means a Type A sequence is inserted into a Type B sequence and separate the Type B sequence into two parts.
In the meantime, to obtain the performance of VSR algorithms if the temporal information can be correctly propagated between the two Type B sequences,
we remove the Type A sequence and combine the two Type B sequences as one~(B-B type in Fig.~\ref{fig:BAB&BB}-(b)).
Then, we apply the BasicVSR on the B-A-B and B-B patch sequence and calculate the PSNR values only on the patches in the Type B sequences.
According to the experiment, the BasicVSR perform much better on the B-B patch sequences (33.74 dB) than on the B-A-B patch sequences (32.40 dB),
which demonstrates that the patches with temporal redundancy in the video sequence will hinder the propagation and cause generalization problem.

\begin{figure*}[!h] 
  \centering
  \includegraphics[width=0.9\linewidth]{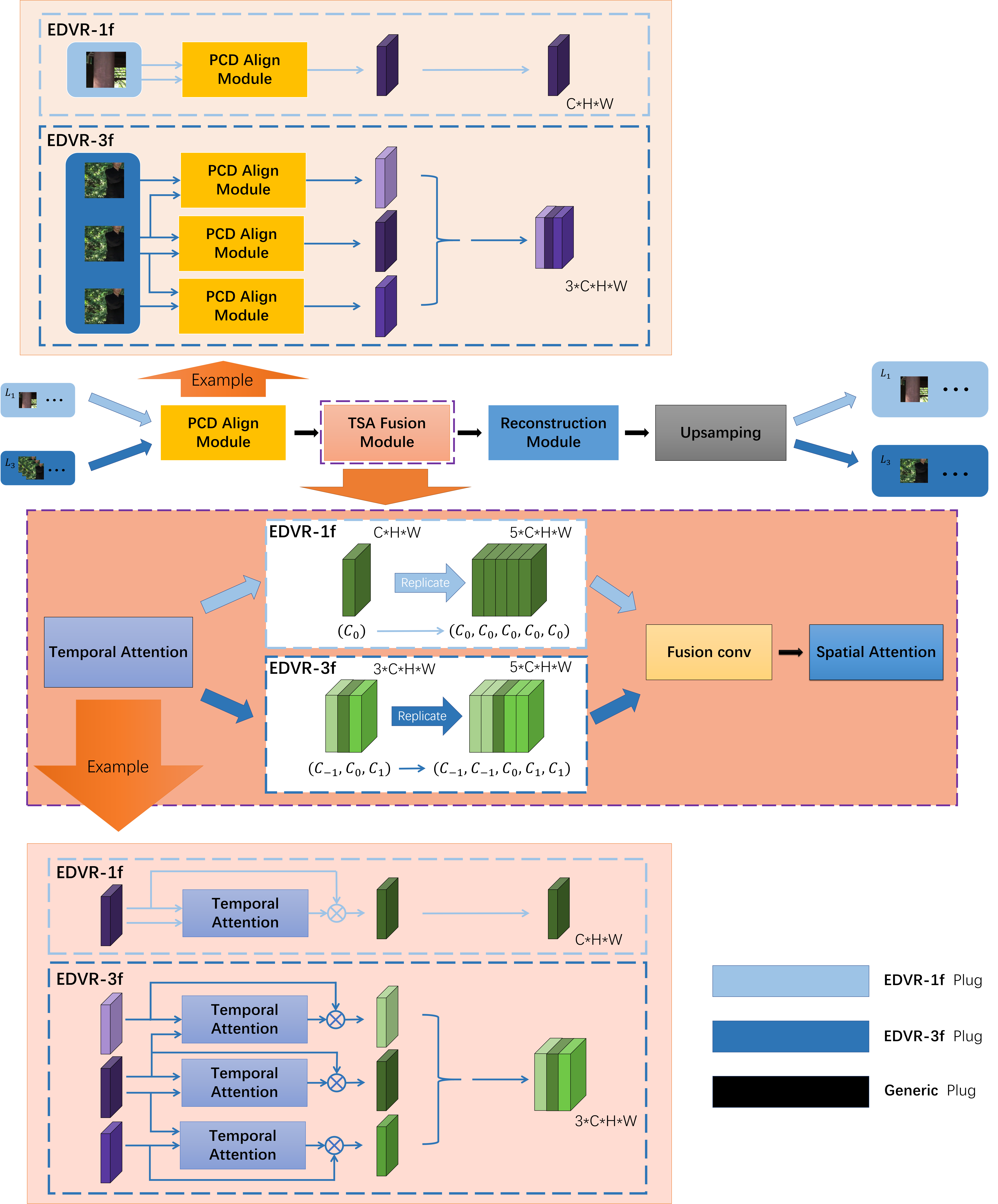}
  \caption{\textbf{Details of the proposed EDVR-1f and EDVR-3f.} 
  }
\label{fig:EDVR-1f&EDVR1-3f}
\end{figure*}

\section{More Details of EDVR-1f and EDVR-3f}
\label{sec:3}

To obtain more suitable models for patch sequences with different motion states,
We slightly modify the EDVR-5f~\cite{wang2019edvr} to acquire the EDVR-1f and EDVR-3f models for single-frame and three frames inputs, respectively.
As shown in Fig.~\ref{fig:EDVR-1f&EDVR1-3f}, unlike the original EDVR-5f, EDVR-1f and EDVR-3f only apply the PCD alignment module and temporal attention in the TSA module for 1 and 3 times, respectively.
Then we replicate the feature in channel dimension to meet the requirement of the subsequent fusion convolution layer in TSA module. 
Specifically, the features in EDVR-1f and EDVR-3f, $(C_{0})$ and $(C_{-1}, C_{0}, C_{1})$, will be replicated to the same shape as the corresponding features in EDVR-5f, i.e., $(C_{0}, C_{0}, C_{0}, C_{0}, C_{0})$ and $(C_{-1}, C_{-1}, C_{0}, C_{1}, C_{1})$, before feeding to the subsequent fusion convolution layer.

\begin{figure*}[!h] 
  \centering
  \includegraphics[width=0.9\linewidth]{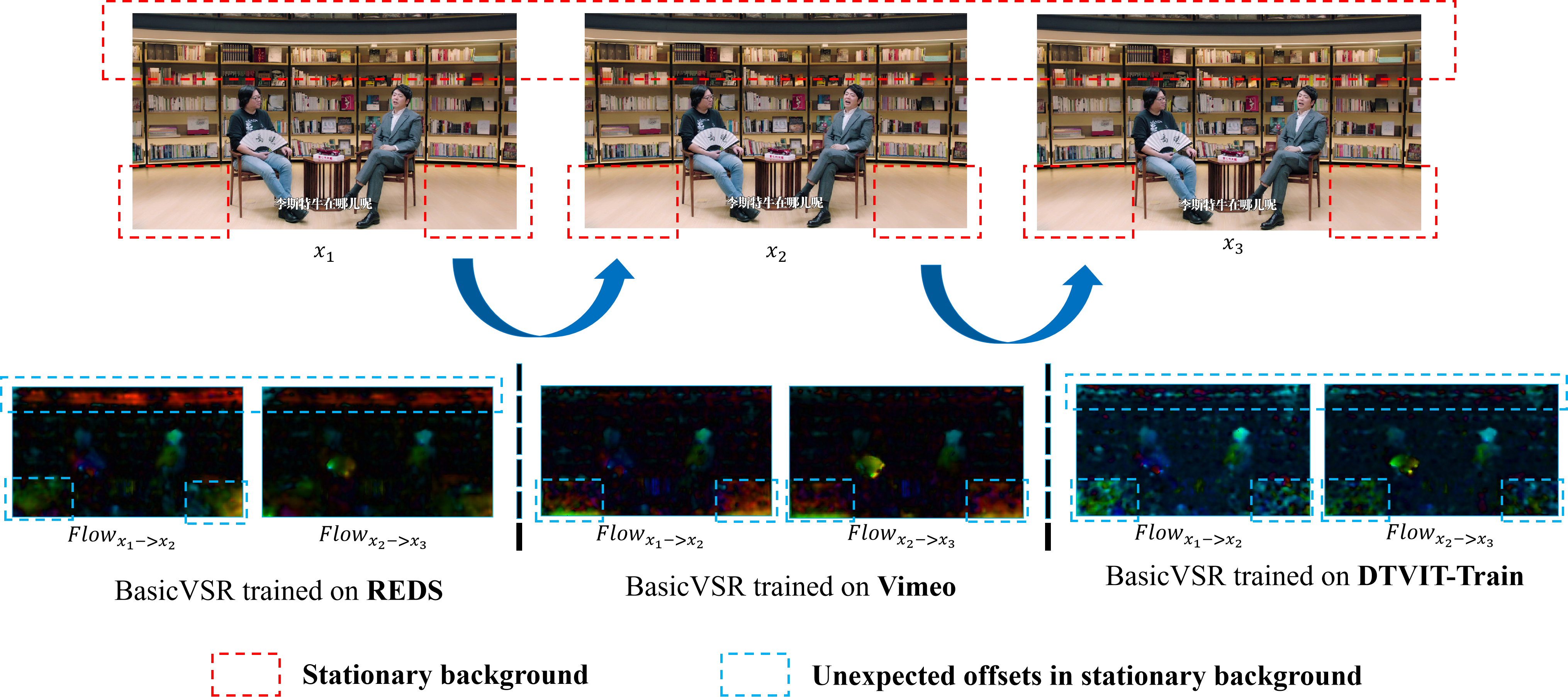}
  \caption{\textbf{Visualization of the estimated optical flow in BasicVSR.} 
  }
\label{fig:Flow Error}
\end{figure*}

\section{Analysis on the Generalization Problem of BasicVSR}
\label{sec:6}
To verify the inferior performance of BasicVSR on DTVIT dataset is caused by the error accumulation of the estimated optical flow~\cite{spy}, we present the intermediate optical flow during the inference time.

As shown in Fig.~\ref{fig:Flow Error}, where we visualize the estimated optical flow between neighboring frames, unexpected offsets occurs in stationary background.
It is also noted that, no matter trained on the REDS datasets or the datasets of similar distribution (e.g., Vimoe and DTVIT-Train datasets), all the BasicVSR models suffer from the inaccurate flows due to regarding these changed pixels due to noisy and information loss during encoding and decoding as the useful temporal information.
Then, the inaccurate flows will be used to perform back-warping on the hidden states and the errors will be pass to next frame via the propagation scheme.
Therefore, the error of the optical flow will be accumulated progressively and cause unsatisfactory super-resolving results.

\begin{figure*}[!h] 
  \centering
  \includegraphics[width=0.8\linewidth]{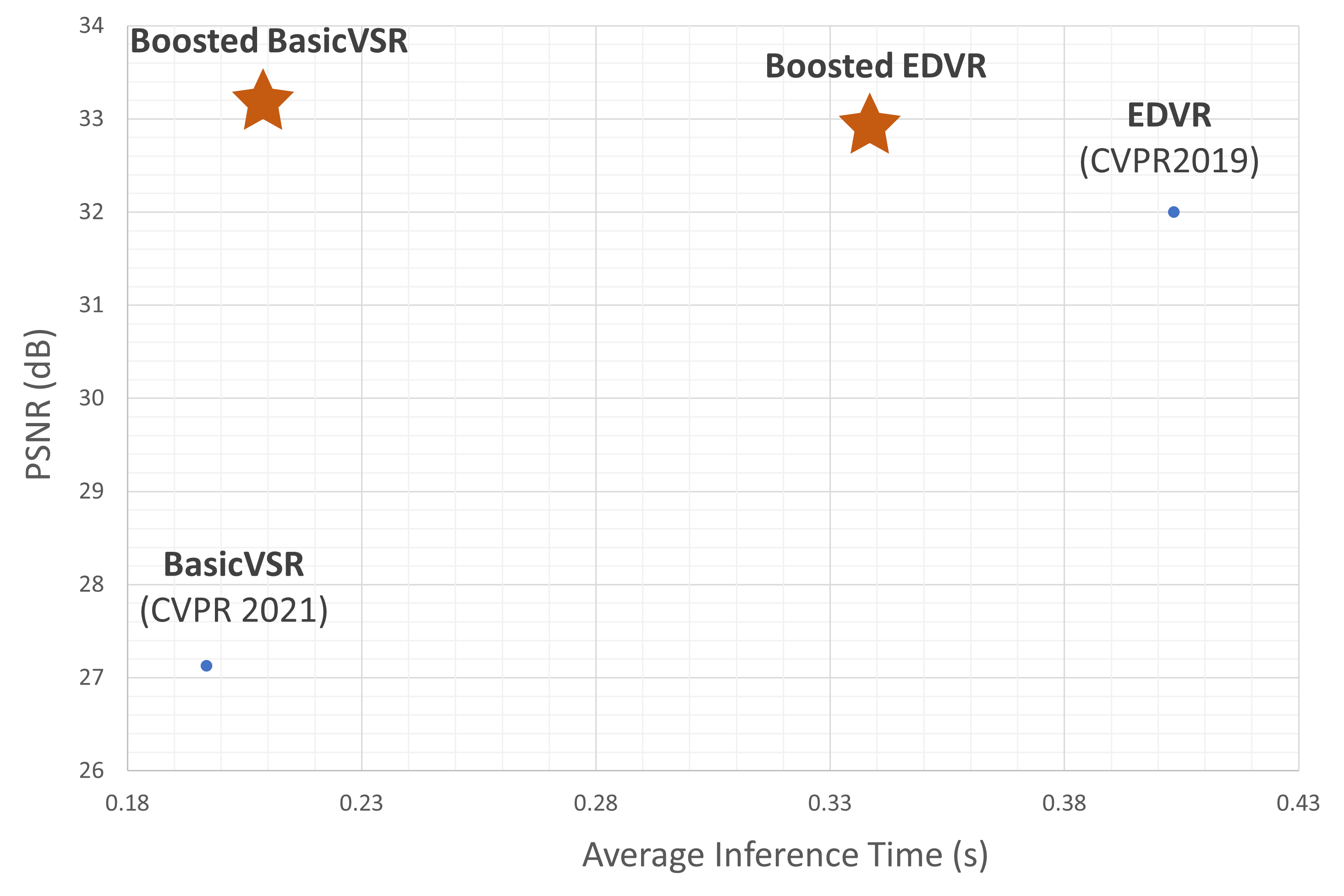}
  \caption{\textbf{Speed and performance comparison.} All the models are trained on the REDS~\cite{REDS} dataset and tested on the DTVIT dataset.
  }
\label{fig:Speed}
\end{figure*}

\section{Efficiency Evaluations}
\label{sec:5}

To verify the efficiency of the proposed method, we evaluate the inference time of the proposed method on a TITAN RTX GPU.
As shown in Fig.~\ref{fig:Speed}, due to more efficient and effective models, EDVR-1f and EDVR-3f, are chosen to super-resolve the patch sequence with stationary objects and background, the proposed \textbf{Boosted EDVR} can achieve better results with less inference time than EDVR.
Although \textbf{Boosted BasicVSR} takes little more time than BasicVSR due to decomposed overlapping patches, it still faster than the the proposed \textbf{Boosted EDVR} and EDVR.

\section{More Quantitative Results}
\label{sec:4}

All the models in the manuscripts are trained on the REDS dataset~\cite{REDS}, which is a first-person video dataset with consistent movement.
To verify the proposed method can not only solve the generalization problem but also can enhance the effectiveness of the propagation branches, we apply the proposed Patch-based Dynamic Propagation (PDP) branch to the BasicVSR trained on the DTVIT-Train to see whether the improvement can be obtained.
As shown in Table~\ref{tab:Vimon_model performance}, the BasicVSR trained on the DTVIT-Train will not suffer of generalization problem and achieves favorable on the DTVIT dataset.
In the meanwhile, the proposed \textbf{Boosted BasicVSR} can still outperform the BasicVSR without any training process, which demonstrates that the PDP can effectively improve the performance of existing non-local propagation-based VSR algorithms.
It is noted that 0.37 dB performance gain based on a SotA method is remarkable for a plug-and-play method without taking more inference time.

\begin{table}
  \centering
  \caption{\textbf{The performance of BasicVSR and Boosted BasicVSR trained on the DTVIT-Train dataset. } }
  \begin{adjustbox}{width=0.7\linewidth}
  \begin{tabular}{l c }
      \hline
      {\bf Method}               & Average PSNR on DTVIT     \\
      \hline
       BasicVSR                  & 33.36 dB   \\ 
       Boosted BasicVSR          & 33.73 dB  \\ 
      \hline  
  \end{tabular}
  \end{adjustbox}
  \label{tab:Vimon_model performance}
  \end{table}

\begin{figure*}[!h] 
  \centering
  \includegraphics[width=0.7\linewidth]{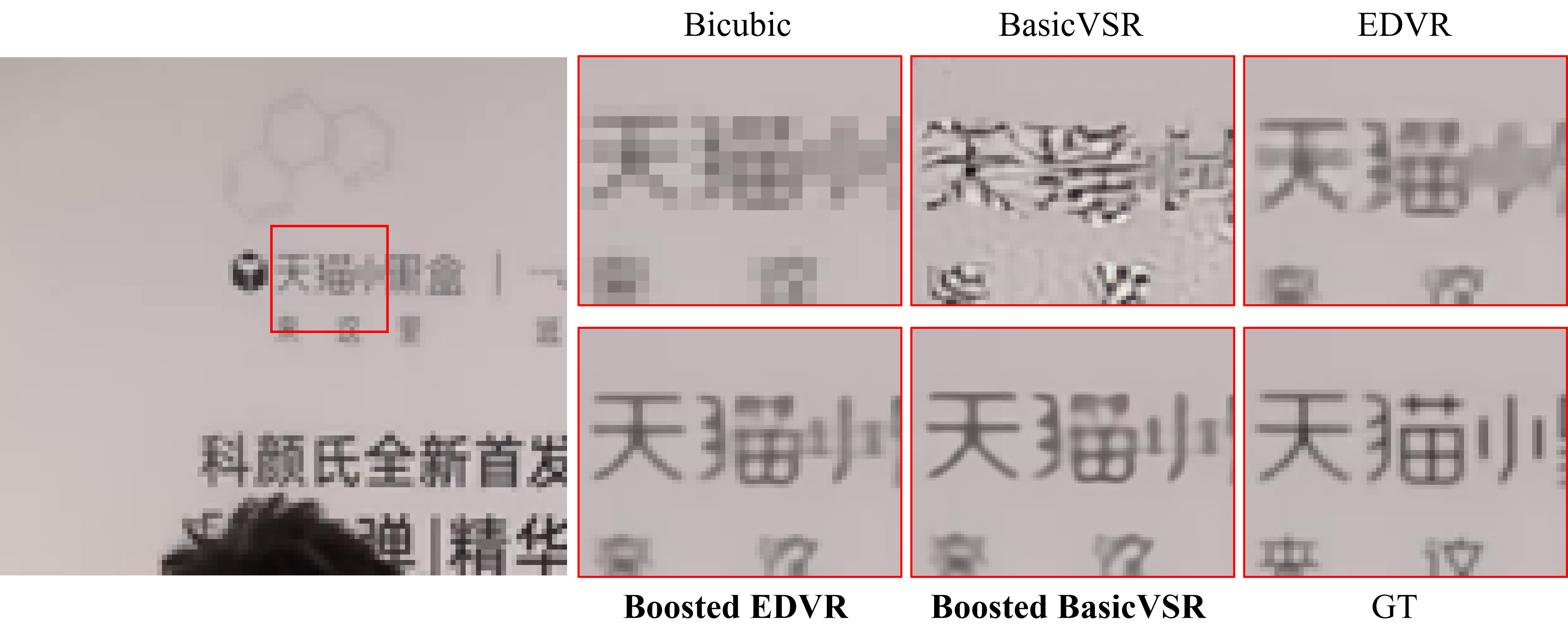}
  \caption{\textbf{Qualitative results of the live streaming scene on the DTVIT dataset.} Best viewed on a high-resolution display.
  }
\label{fig:Live_streaming}
\end{figure*}

\begin{figure*}[!h]  
  \centering
  \includegraphics[width=0.7\linewidth]{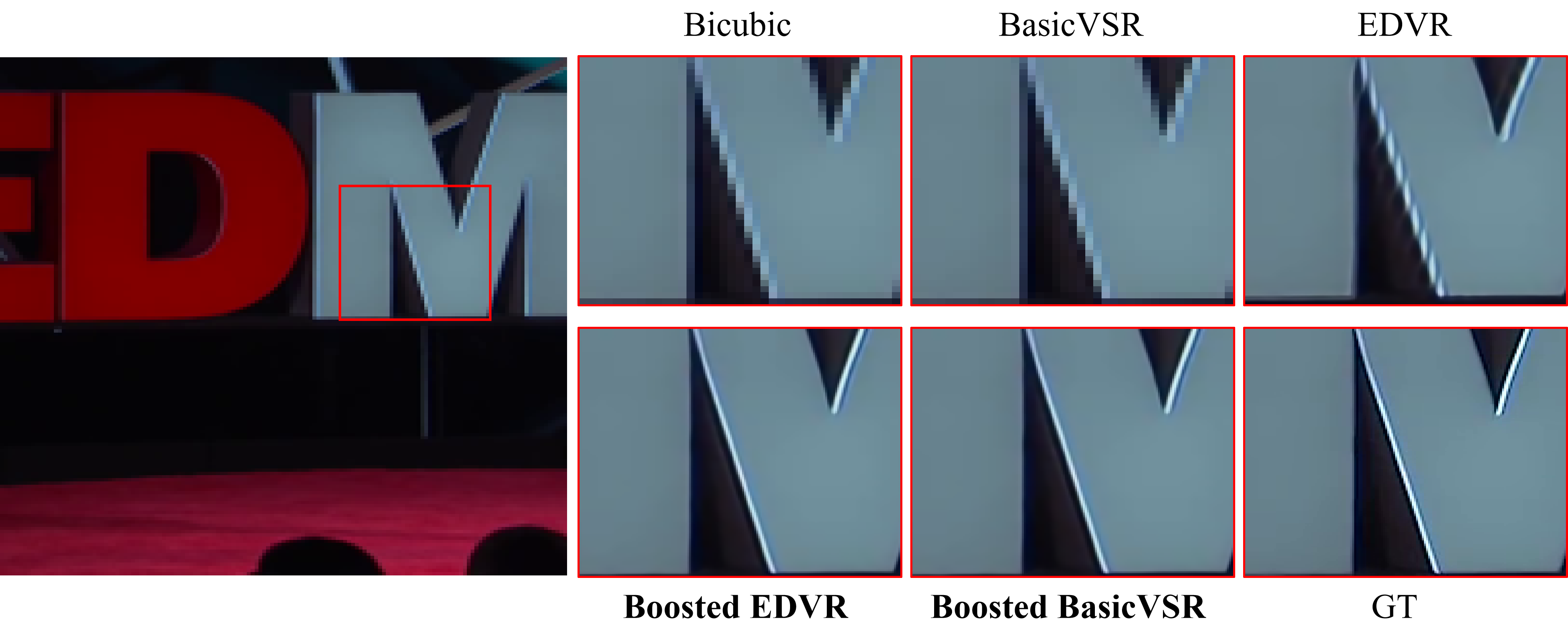}
  \caption{\textbf{Qualitative results of the TV program scene on the DTVIT dataset.} Best viewed on a high-resolution display.
  }
\label{fig:TV_program}
\end{figure*}

\begin{figure*}[!h] 
  \centering
  \includegraphics[width=0.7\linewidth]{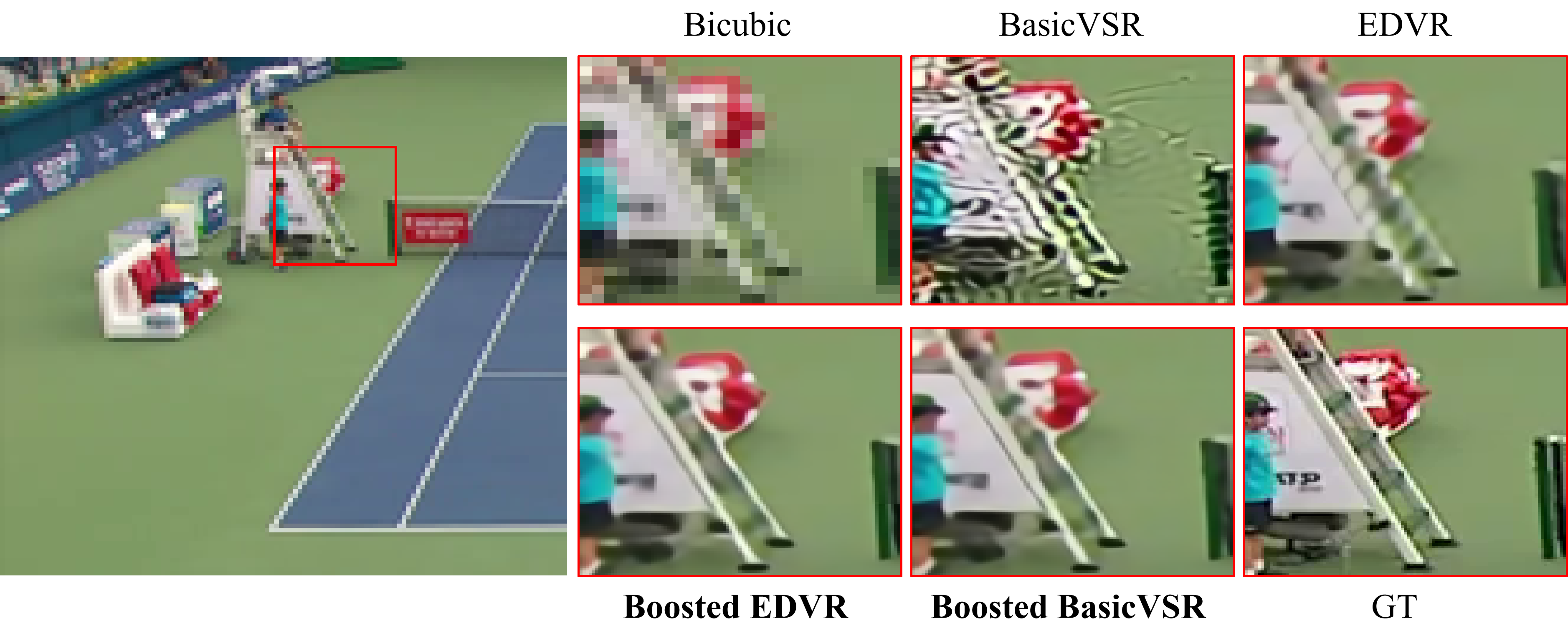}
  \caption{\textbf{Qualitative results of the sports live scene on the DTVIT dataset.} Best viewed on a high-resolution display.
  }
\label{fig:Sports_live}
\end{figure*}

\begin{figure*}[!h] 
  \centering
  \includegraphics[width=0.7\linewidth]{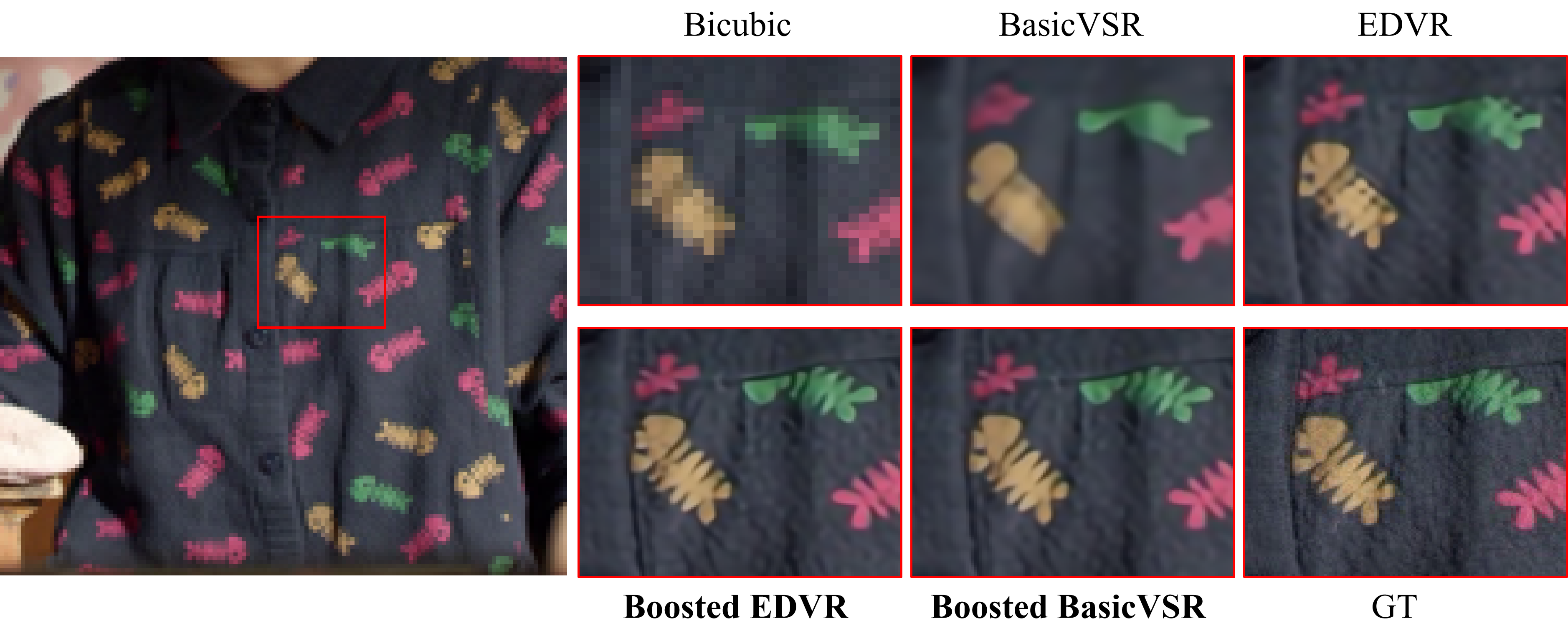}
  \caption{\textbf{Qualitative results of the movie and television play scene on the DTVIT dataset.} Best viewed on a high-resolution display.
  }
\label{fig:Movie_and_television_play}
\end{figure*}

\begin{figure*}[!h]  
  \centering
  \includegraphics[width=0.7\linewidth]{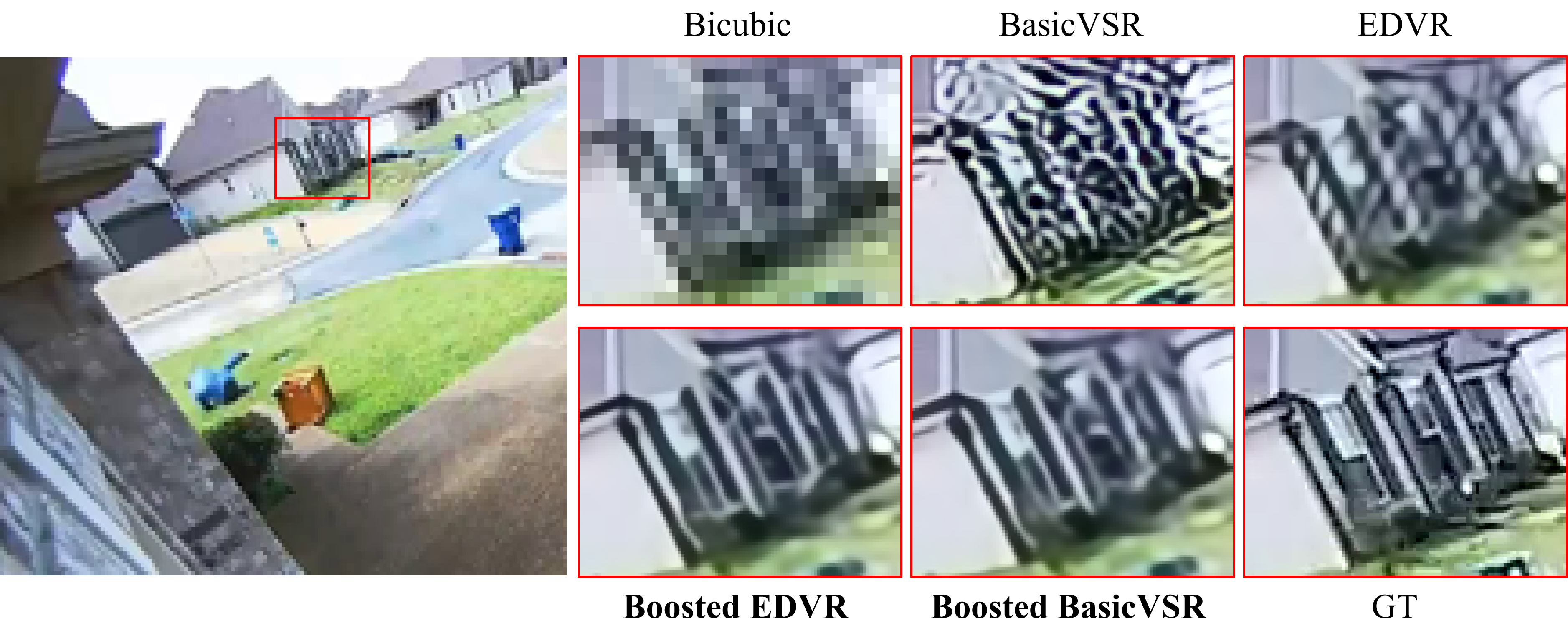}
  \caption{\textbf{Qualitative results of the surveillance video scene on the DTVIT dataset.} Best viewed on a high-resolution display.
  }
\label{fig:Surveillance_video}
\end{figure*}

\begin{figure*}[!h] 
  \centering
  \includegraphics[width=0.7\linewidth]{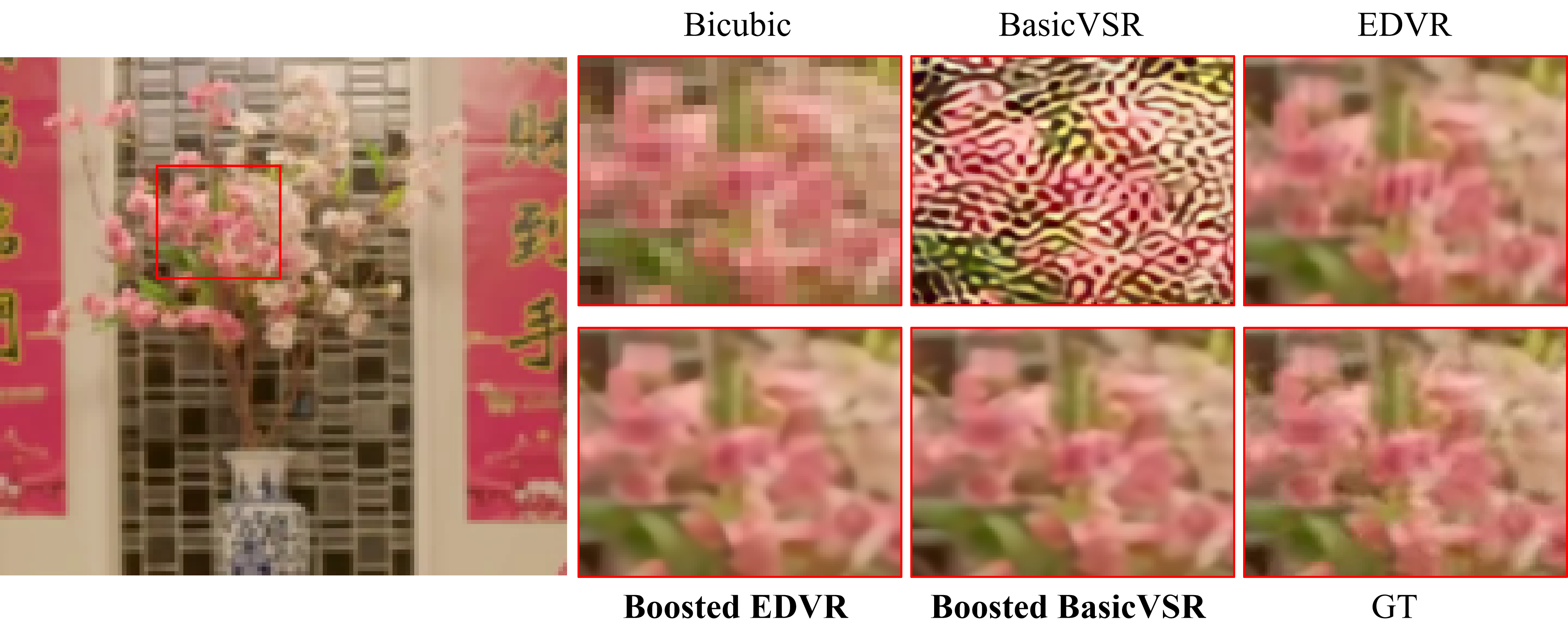}
  \caption{\textbf{Qualitative results of the advertisement scene on the DTVIT dataset.} Best viewed on a high-resolution display.
  }
\label{fig:Advertisement}
\end{figure*}

\begin{figure*}[!h]  
  \centering
  \includegraphics[width=0.7\linewidth]{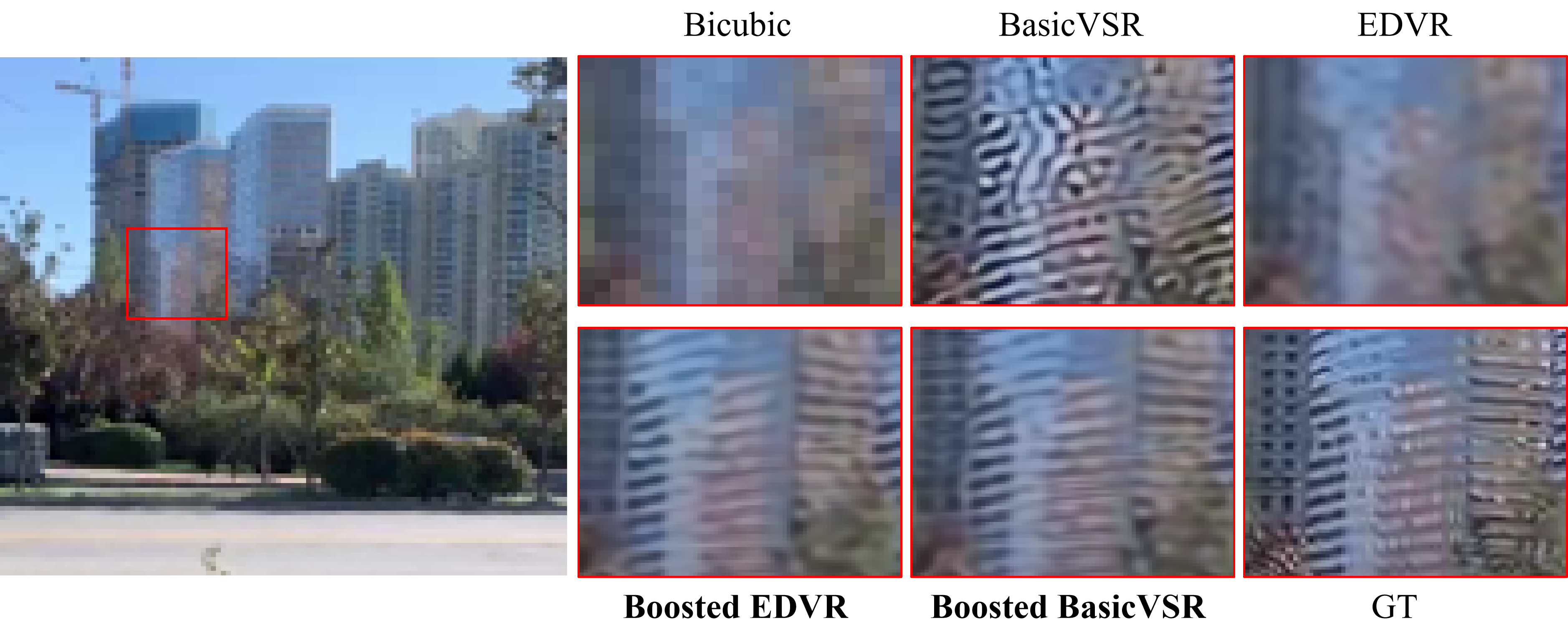}
  \caption{\textbf{Qualitative results of the first-person video scene on the DTVIT dataset.} Best viewed on a high-resolution display.
  }
\label{fig:First_person_videos}
\end{figure*}

\begin{figure*}[!h] 
  \centering
  \includegraphics[width=0.6\linewidth]{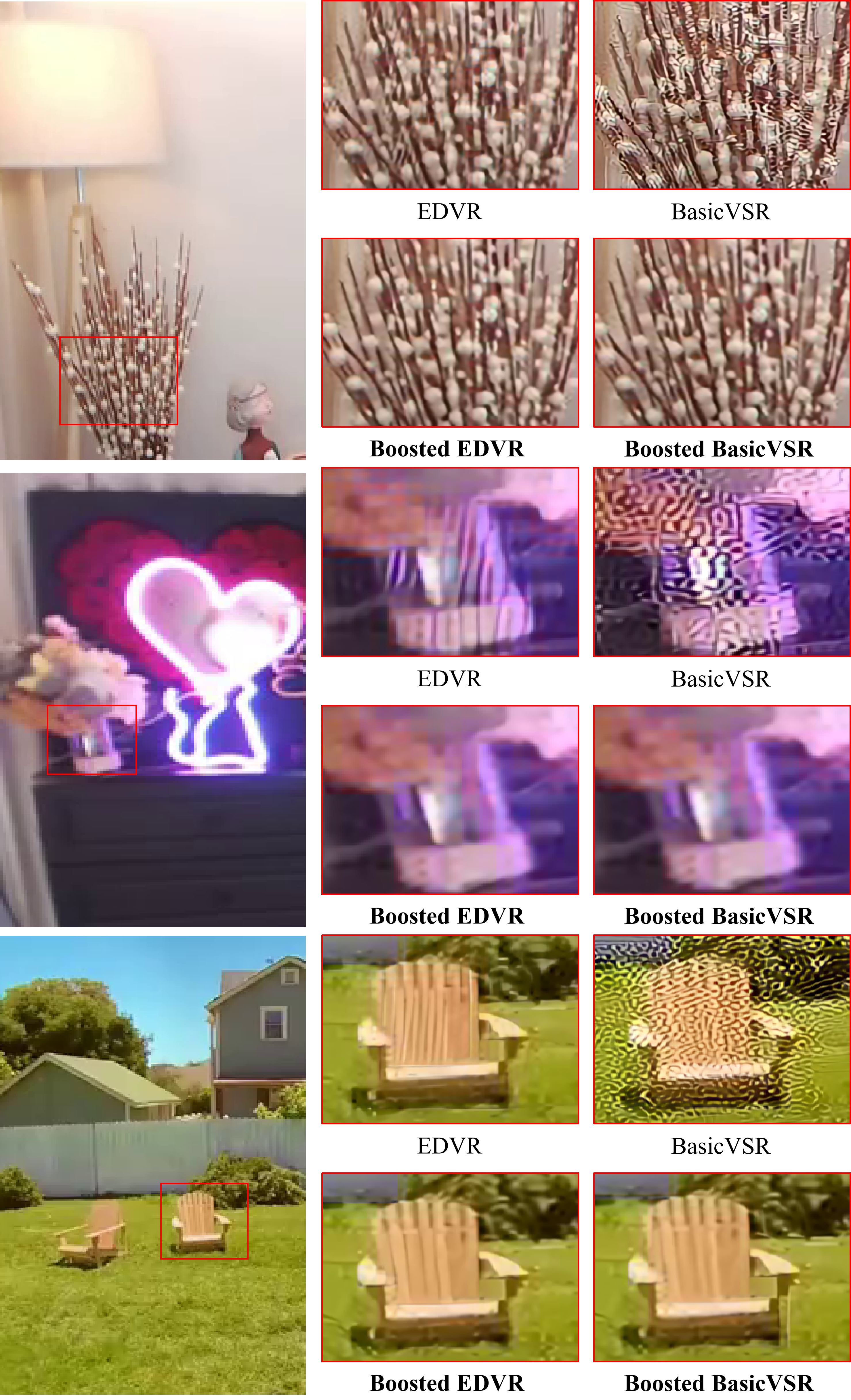}
  \caption{\textbf{Qualitative comparison on the real-world videos.} Best viewed on a high-resolution display.
  }
\label{fig:Real-World Video Qualitative comparison}
\end{figure*}

\section{More Qualitative Results}
\label{sec:7}

In this section, we provide additional qualitative results to clearly show the effectiveness of the proposed method. 
From Fig.~\ref{fig:Live_streaming} to Fig.~\ref{fig:First_person_videos}, we select one scene from each type of the DTVIT dataset and show the 
super-resolving results of different methods.
Besides, we also collect some real-world videos with low-quality and low-resolution to evaluate the performance of the state-of-the-art VSR algorithms on realistic scenes~(Fig.~\ref{fig:Real-World Video Qualitative comparison}).
For the figures from the DTVIT dataset, we present the results from bicubic upsampling, EDVR and BasicVSR trained on the REDS dataset, the proposed \textbf{Boosted BasicVSR} and \textbf{Boosted EDVR}.
For the figures from the real-world videos, we only present the results from EDVR, BasicVSR, \textbf{Boosted BasicVSR}, and \textbf{Boosted EDVR}.

All the qualitative comparisons demonstrate that the proposed \textbf{Boosted BasicVSR} and \textbf{Boosted EDVR} are able to reconstruct images with more details by effectively exploiting the temporal information.

\bibliography{aaai23}